\documentclass[runningheads]{llncs}
\usepackage{graphicx}
\usepackage[dvipsnames]{xcolor}
\usepackage{tikz}
\usepackage{comment}
\usepackage{epsfig}
\usepackage{graphicx}
\usepackage{amsmath,amssymb} % define this before the line numbering.
\usepackage{multirow}
\usepackage{color}
\usepackage{siunitx}
\usepackage{ulem}

\usepackage[width=132mm,left=15mm,paperwidth=162mm,height=193mm,top=12mm,paperheight=217mm]{geometry}

\newlength\savewidth
\definecolor{lightblue}{HTML}{1155CC}
\newcommand{\lidar}{LiDAR}
\newcommand{\radar}{Radar}

\begin{document}
\pagestyle{headings}
\mainmatter
\def\ECCVSubNumber{3017}  % Insert your submission number here

\title{RadarNet: Exploiting \radar\ for Robust Perception of Dynamic Objects} % Replace with your title

\begin{comment}
\titlerunning{ECCV-20 submission ID \ECCVSubNumber} 
\authorrunning{ECCV-20 submission ID \ECCVSubNumber} 
\author{Anonymous ECCV submission}
\institute{Paper ID \ECCVSubNumber}
\end{comment}
%******************

%\begin{comment}
\titlerunning{RadarNet: Exploiting \radar\ for Robust Perception of Dynamic Objects}
\author{Bin Yang\inst{1,2}\thanks{Equal contribution. Work done during RG's internship at Uber ATG.},
Runsheng Guo\inst{3}$^{\star}$,
Ming Liang\inst{1},
Sergio Casas\inst{1,2},\\
Raquel Urtasun\inst{1,2}}
\authorrunning{B. Yang et al.}
\institute{$^1$Uber Advanced Technologies Group,
$^2$Univeristy of Toronto,
$^3$University of Waterloo\\
\email{\{byang10,ming.liang,sergio.casas,urtasun\}@uber.com, r9guo@edu.uwaterloo.ca}}
%\end{comment}
%******************
\maketitle

% !TEX root = top.tex
\begin{abstract}
We tackle the problem of exploiting \radar\ for perception in the context of self-driving as \radar\ provides complementary information to other sensors such as \lidar\ or cameras in the form of Doppler velocity.
The main challenges of using  \radar\ are the noise and measurement ambiguities  which have been a struggle for existing simple input or output fusion methods.
To better address this, we propose a new solution that exploits both \lidar\ and \radar\ sensors for perception.
Our approach, dubbed RadarNet, features a voxel-based early fusion and an attention-based late fusion, which learn from data to exploit both geometric and dynamic information of \radar\ data.
RadarNet achieves state-of-the-art results on two large-scale real-world datasets in the tasks of object detection and velocity estimation.
We further show that exploiting \radar\ improves the perception capabilities of detecting faraway objects and understanding the motion of dynamic objects.
\end{abstract}

% !TEX root = top.tex
\section{Introduction}

Self-driving vehicles (SDVs) have to perceive the world around them in order to interact with the environment in a safe manner.
Perception systems typically detect the objects of interest and track them over time in order to estimate their motion.
Despite many decades of research, perception systems have not achieved the level of reliability required to deploy self-driving vehicles at scale without safety drivers.

Recent 3D perception systems typically exploit cameras \cite{monodis,mono3d,weng2019monocular}, \lidar~\cite{pixor,shi2019pointrcnn,lang2019pointpillars}, or their combination \cite{fpointnet,contfuse,mv3d} to achieve high-quality 3D object detection. While cameras capture rich appearance features, \lidar\ provides direct and accurate 3D measurements. The sparsity of \lidar\ measurements (e.g., at long range) and the sensor's sensitivity to weather (e.g., fog, rain and snow) remain open challenges.
In addition to detecting and recognizing objects, estimating their velocities is also of vital importance. In some safety critical situations, for example a child running out of occlusion in front of the SDV, the SDV needs to estimate velocities from a single measurement cycle in order to avoid collision.
This estimation is often inaccurate (or even impossible) when using \lidar\ or cameras alone as they provide static information only.
While for pedestrians we may infer the motion from its pose with large uncertainty, for rigid objects like vehicles we can not make reasonable predictions from their appearance alone.

An appealing solution is to use sensors that are robust to various weather conditions and can provide velocity estimations from a single measurement.
This is the case of \radar, which uses the Doppler effect to compute the radial velocities of objects relative to the SDV.
\radar\ brings its own challenges, as the data is very sparse (typically much more so than \lidar), the measurements are ambiguous in terms of  position and velocity, the readings lack tangential information and often contain false positives.
As a result, previous methods either focus on the ADAS by fusing \radar\ with cameras \cite{chadwick2019distant,nobis2019deep,nabati2019rrpn,danzer20192d}, where the performance requirements are relatively low; or fuse \radar\ data at the perception output level (e.g., tracks) \cite{cho2014multi,hajri2018real,gohring2011radar}, thus failing to fully exploit the complementary information of the sensors.

In this paper, we take a step forward in this direction and design a novel neural network architecture, dubbed {\it RadarNet},  which can exploit both \lidar\ and \radar\ to provide accurate detections and velocity estimates for the actors in the scene.
Towards this goal, we propose a multi-level fusion scheme that can fully exploit both geometric and dynamic information of \radar\ data.
In particular, we first fuse \radar\ data with \lidar\ point clouds via a novel {\it voxel-based early fusion} approach to leverage the \radar's long sensing range.
Furthermore, after we get object detections, we fuse \radar\ data again via an {\it attention-based late fusion} approach to leverage the \radar's velocity readings.
The proposed attention module captures the uncertainties in both detections and \radar\ measurements and plays an important role in transforming the 1D radial velocities from \radar\ to accurate 2D object velocity estimates.

We demonstrate the effectiveness of RadarNet on two large-scale driving datasets, where it surpasses the previous state-of-the-art in both 3D object detection and velocity estimation. We further show that exploiting \radar\ brings significant improvements in perceiving dynamic objects, improving both motion estimation and long range detection.
% !TEX root = top.tex
\section{Related Work}

\subsubsection{Exploiting \lidar\ for Perception:}
As a high-quality 3D sensor, \lidar\ has been widely used for 3D object detection in self-driving. Previous methods mainly differ in two aspects: the detection architecture and the input representation. While single-stage detectors \cite{voxelnet,pixor,lang2019pointpillars} have the advantages of simplicity and fast inference, two-stage methods \cite{mv3d,avod,shi2019pointrcnn} are often superior in producing precisely localized bounding boxes.
Different representations of \lidar\ point clouds have been proposed: 3D voxel grids \cite{li20173d}, range view (RV) projections \cite{li2016vehicle,meyer2019lasernet,mv3d}, bird's eye view (BEV) projections \cite{voxelnet,pixor,second}, and point sets \cite{shi2019pointrcnn,fpointnet,shi2020point} are amongst the most popular. While 3D voxel grids are slow and wasteful to process due to the size of the volume which is mainly sparse, range view projections are dense representations by nature. However, RV images suffer from the large variance in object size and shape due to the projection. BEV projections achieve a better trade-off between accuracy and speed. Voxel features represented with either simple statistics \cite{pixor,mv3d} or learned representations \cite{lang2019pointpillars} have been proposed.
In this paper, we use a single-stage detector with BEV representation for its simplicity, effectiveness and efficiency.

\subsubsection{Exploiting \radar\ for Perception:}
\radar\ has long been used in ADAS for adaptive cruise control and collision avoidance due to its  cost and robustness to severe weather conditions.
Recently, \radar\ has been exploited in many other applications, spanning across free space estimation \cite{sless2019road,lombacher2017semantic}, object detection \cite{chadwick2019distant,nabati2019rrpn,danzer20192d}, object classification \cite{kim2018moving,wohler2017comparison,patel2019deep} and segmentation \cite{sless2019road,schumann2018semantic,lombacher2017semantic}.
However, most of these methods treat \radar\ as another 3D sensor, ignoring its high-fidelity velocity information. In contrast, we exploit both \radar's geometric and dynamic information  thanks to  a novel specialized fusion mechanism for each type of information.

\begin{figure}[t]
\centering
\includegraphics[width=1.0\linewidth]{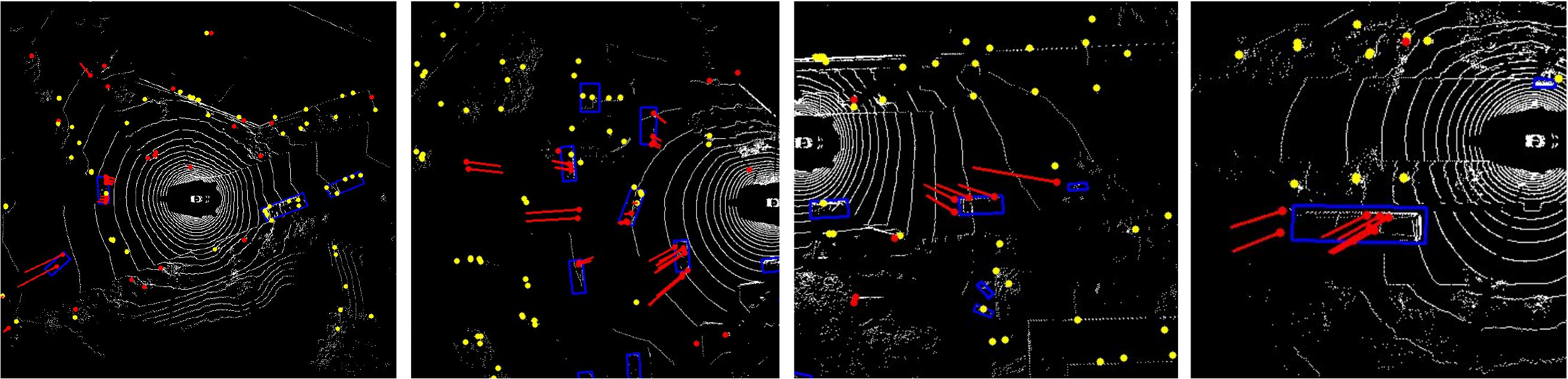}
\caption{{\bf \lidar\ and \radar\ sensor data:} We show \lidar\ data in white, dynamic \radar\ returns (with radial velocity) in {\color{red}red}, static \radar\ returns in {\color{Goldenrod}yellow}, and object labels in {\color{blue}blue}.}
\label{fig:sensor_vis}
\end{figure}

\subsubsection{Sensor Fusion with \radar:}
In many self-driving perception systems, \radar\ data has been fused at the perception output level in the form of object tracks \cite{cho2014multi,hajri2018real,gohring2011radar}. Kalman Filter \cite{sun2004multi} or IMM \cite{blom1988interacting} trackers are popular approaches  to digest \radar\ data, and the resulting tracks are then fused with object tracks from other sensors. However, sensor fusion is not exploited during the process of generating those object tracks.
Recent works also look at fusion between \radar\ and cameras within the perception system. Different \radar\ representations are proposed to facilitate fusion: spectrogram images \cite{limradar}, sparse locations in image space \cite{nabati2019rrpn}, pseudo-image by projecting to image space \cite{nobis2019deep,chadwick2019distant}, BEV representation \cite{meyer2019deep} and object detections \cite{kuang2020multi}.
However, these methods do not have high accuracy in terms of  3D perception.
Instead, here we choose to fuse \radar\ with \lidar\, and design a multi-level fusion mechanism that outperforms the state-of-the-art in self-driving.
%!TEX root = top.tex
\section{Review of \lidar\ and \radar\ Sensors}

% !TEX root = ../top.tex

\setlength{\tabcolsep}{4pt}
\begin{table}[t]
\begin{center}
\caption{{\bf Hardware comparison between \lidar\ and \radar\ sensors}}
\label{tab:sensor_accuracy}
\begin{tabular}{lcccc}
\hline\noalign{\smallskip}
Sensor & Detection & Range & Azimuth & Velocity \\
Modality & Range   & Accuracy & Resolution & Accuracy \\
\noalign{\smallskip}
\hline
\noalign{\smallskip}
\lidar\ & 100 m & 2 cm & \ang{0.1} $\sim$ \ang{0.4} & - \\
\multirow{2}{*}{\radar} & \multirow{2}{*}{250 m} & 10 cm near range & \ang{3.2} $\sim$ \ang{12.3} near range & \multirow{2}{*}{0.1 km/h}\\
                       &                        & 40 cm far range & \ang{1.6} far range & \\
\hline
\end{tabular}
\end{center}
\end{table}

We first provide a review of \lidar\ and \radar\ sensors and introduce our notation. We hope this short review can help readers better understand the intuitions behind our model designs, which will be described in the next section.

\lidar\ (light detection and ranging) sensors can be divided into three main types: spinning \lidar, solid state \lidar, and flash \lidar. In this paper we focus on the most common type: spinning \lidar. This type of  \lidar\, emits and receives laser light pulses in \ang{360} and exploits the time of flight (ToF)  to calculate the distance to the obstacles.
As a result, \lidar\ data is generated as a continuous stream of point clouds. We denote each \lidar\ point as a vector $P=(x,y,z,t)$, encoding  the 3D position and the capture timestamp.
In practice we often divide the \lidar\ data into consecutive \ang{360} sweeps for frame-wise point cloud processing.
\lidar\ is the preferred sensor for most self-driving vehicles due to its accurate 3D measurements.
The main drawbacks are its sensitivity to dirt (which leads to poor performance in fog, rain and snow), cold (that causes exhaust plumes) as well as the lack of reflectivity of certain materials (such as windows and certain paints). Furthermore, its density decreases with range, making long range detection challenging.

\radar\ (radio detection and ranging) sensors work similarly as \lidar, but transmit electromagnetic waves to sense the environment.
The \radar\ outputs can be organized in three different levels: raw data in the form of time-frequency spectrograms, clusters from applying DBSCAN \cite{kellner2012grid} or CFAR \cite{skolnik1990Radar} on raw data, and tracks from performing object tracking on the clusters. From one representation to the next, the data sparsity and abstraction increases, while the noise in the data decreases. In this paper we focus on the mid-level data form, \radar\ clusters, for its good balance between information richness and noise.
In the following we refer to these clusters as \radar\ targets. We denote each \radar\ target as a vector $Q=(\mathbf{q},v_\|,m,t)$, where $\mathbf{q}=(x, y)$ is the 2D position in BEV, $v_\|$ is a scalar value representing the radial velocity, $m$ is a binary value indicating whether the target is moving or not, and $t$ is the capture timestamp.
The main advantages of \radar\ are that it provides instantaneous velocity measurements and is robust to various weather conditions.
However, its drawbacks are also significant. It has a low resolution and thus it is difficult to detect small objects. There are ambiguities (a modulo function) in range and velocity due to \radar\ aliasing, as well as false positive detections from clutter and multi-path returns.
It is also worth noting that the objects' real-world velocities (2D vectors in BEV) are ambiguous given only the radial velocity. Therefore we need to additionally estimate the tangential velocity or the 2D velocity direction in order to properly utilize the radial velocity.

We compare \lidar\ and \radar\ data both quantitatively and qualitatively.
We visualize both sensors' data from the nuScenes dataset \cite{caesar2019nuscenes} in Fig.~\ref{fig:sensor_vis}, and we compare their technical specifications in Table~\ref{tab:sensor_accuracy}.
Note that \lidar\ outperforms \radar\ in both accuracy and resolution by over an order of magnitude. The accurate 3D surface measurements makes \lidar\ the first choice for high-precision 3D object detection. \radar\ can provide complementary information in two aspects: more observations at long range and instantaneous velocity evidence from the Doppler effect.
We thus argue that since these sensors are very complementary, their combination provides a superior solution for self-driving.

% !TEX root = top.tex
\section{Exploiting \lidar\ and \radar\ for Robust Perception}

\begin{figure}[t]
\centering
\includegraphics[width=1.0\linewidth]{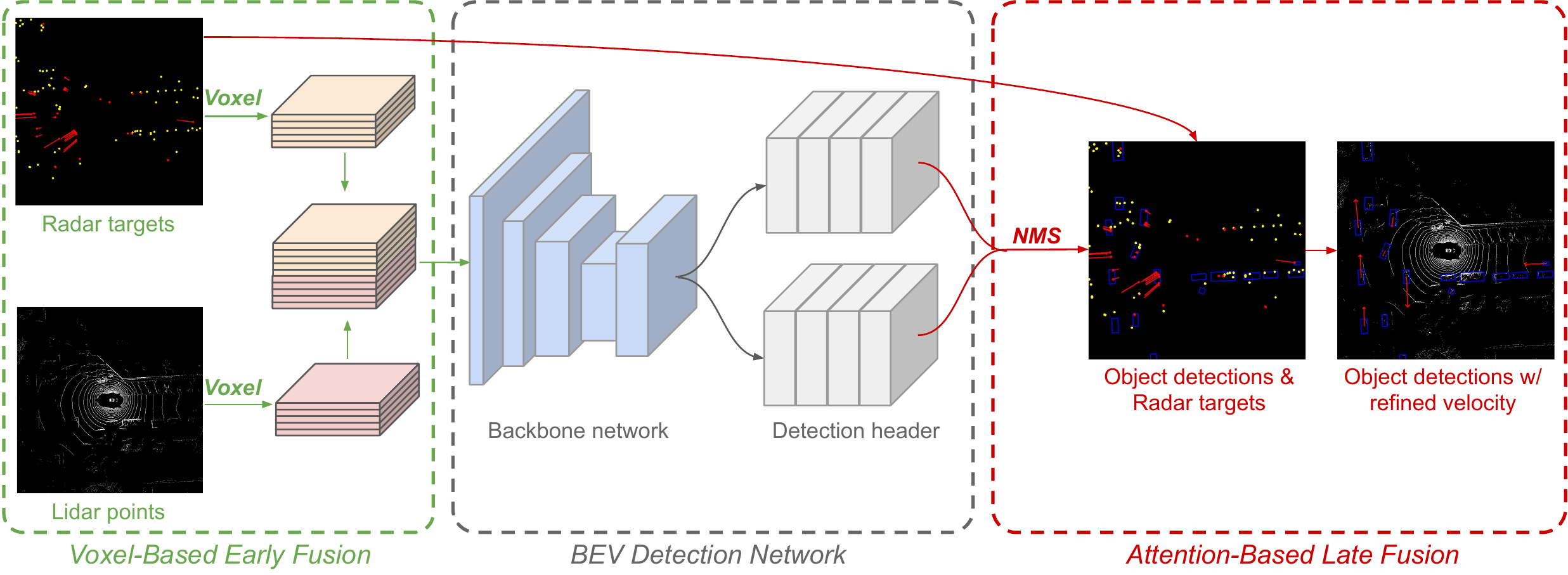}
\caption{{\bf RadarNet:} Multi-level \lidar\ and \radar\ fusion is performed  for accurate 3D object detection and velocity estimation.}
\label{fig:model_arch}
\end{figure}

In this section we present our novel approach to 3D perception, involving 3D object detection and velocity estimation.
We refer the reader to Fig.~\ref{fig:model_arch} for an illustration of the  overall architecture of our approach.
To fully exploit the complementary information of the two sensor modalities and thereby benefit both object detection and velocity estimation, we propose two sensor fusion mechanisms, namely {\it early fusion} and {\it late fusion}, that operate at different granularities.
More specifically, while early fusion learns joint representations from both sensor observations, late fusion refines object velocities via an attention-based association and aggregation mechanism between object detections and Radar targets.

\subsection{Exploiting Geometric Information via Early Fusion}

\subsubsection{\lidar\ Voxel Representation:}
We take multiple sweeps of \lidar\ point clouds (those within the past 0.5 seconds) as input so that the model has enough information to infer the objects' motion while  still being able to run in real-time.
All point cloud sweeps  are transformed to the ego-vehicle's centric coordinates at the current frame. Note that this is easy to do as   sensors are calibrated  and  the vehicle pose is estimated by the localization system.
Following FAF \cite{dpt}, we adopt a bird's eye view (BEV) representation and concatenate multiple height slices and sweeps together along the channel dimension.
We use a weighted occupancy value as each voxel's feature representation. Specifically, for each voxel, if no point falls in it, the voxel's value is 0. If one or more points $\{(x_i,y_i,z_i), i=1\ldots N\}$ fall into it, the voxel's value is defined as $\sum_i(1-\frac{|x_i-a|}{\mathrm{d}x/2})(1-\frac{|y_i-b|}{\mathrm{d}y/2})(1-\frac{|z_i-c|}{\mathrm{d}z/2})$, where $(a,b,c)$ is the voxel's center and $(\mathrm{d}x,\mathrm{d}y,\mathrm{d}z)$ is the voxel's size.

\subsubsection{\radar\ Voxel Representation:}
Similar to how we accumulate multiple sweeps of \lidar\ data, we also take multiple cycles of \radar\ data as input, in the same coordinate system as \lidar. We keep only the $(x,y)$ position of \radar\ targets and ignore the height position as it is often inaccurate (if it ever exists). As a result, each cycle of \radar\ data can be voxelized as one BEV image.  We concatenate multiple cycles along the channel dimension and use a motion-aware occupancy value as the feature for each voxel. Specifically, for each BEV voxel, if no \radar\ target falls into it, the voxel's value is 0. If at least one moving \radar\ target (i.e., $m=1$) falls into it, the voxel's value is 1. If all \radar\ targets falling into it are static, the voxel's value is -1.

\subsubsection{Early Fusion:}
We use the same BEV voxel size for \lidar\ and \radar\ data.
Thus their voxel representations have the same size in BEV space.
We perform early fusion by concatenating them together along the channel dimension.

\subsection{Detection Network}
We adopt a single-stage anchor-free BEV object detector with additional velocity estimation  in the detection header.

\subsubsection{Backbone Network:}
We adopt the same backbone network architecture as PnPNet \cite{pnpnet}. The backbone network is composed of three initial convolution layers, three consecutive multi-scale inception blocks \cite{szegedy2015going}, and a feature pyramid network \cite{fpn}. The three initial convolution layers down-sample the voxel input by 4 and output 64-D feature maps. The inception block consists of three branches, each with a down-sampling ratio of $1\times$, $2\times$ and $4\times$ implemented by stride of the first convolution.
The number of convolution layers in each branch is $2$, $4$ and $6$, and the number of feature channels in each branch is $32$, $64$ and $96$. The feature pyramid network merges multi-scale feature maps from the inception block into one, with $256$ channels for each layer. The final output of the backbone network is a $256$-D feature map with a $4\times$ down-sampling ratio compared to the voxel input.

\subsubsection{Detection Header:}
We apply a fully-convolutional  detection header \cite{focal} for anchor-free dense detection, which consists of a classification branch and a regression branch, each with $4$ convolution layers and $128$ channels. The detection is parameterized as $D = (c, x, y, w, l, \theta, \mathbf{v})$, which represents the confidence score, the object's center position in BEV, its width, length and orientation, and its 2-D velocity $\mathbf{v}=(v_x,v_y)$ in BEV. The classification branch predicts the confidence score $c$, while the regression branch predicts all the other terms $(x-p_x, y-p_y, w, l, \cos(\theta), \sin(\theta), m, v_x, v_y)$, where $p_x$ and $p_y$ are the 2D coordinates of every voxel center and $m$ is an additional term that indicates the probability of moving.
During inference, we set the 2-D velocity to $(0, 0)$ if the predicted probability of moving is smaller than 50\%.

\subsection{Exploiting Dynamic Information via Late Fusion}

\begin{figure}[t]
\centering
\includegraphics[width=1.0\linewidth]{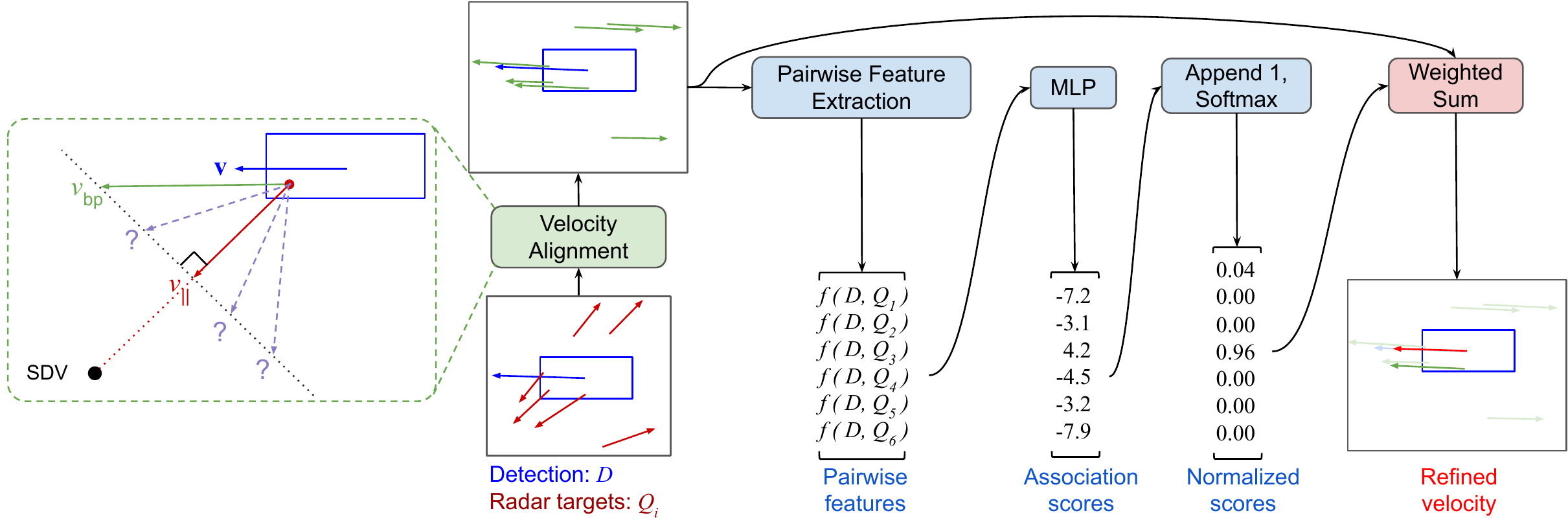}
\caption{{\bf Attention-based late fusion of object detection and Radar targets:} In the figure we show an example of fusing \radar\ with one detection, while in practice this is applied to all detections in parallel. We first {\color{YellowGreen}align} the radial velocities of \radar\ targets with the detection's motion direction, then predict {\color{lightblue}pairwise association scores} for all detection-\radar\ pairs. The {\color{red}refined velocity} is computed as a weighted sum of all \radar\ evidences as well as the original velocity estimate.}
\label{fig:late_fuse}
\end{figure}

While early fusion exploits the position and density information of \radar\ targets, late fusion is designed to explicitly exploit the \radar's radial velocity evidence.
Due to the lack of tangential information, the actual object velocity vector is ambiguous given the radial velocity alone.
To address this issue, we propose to use the velocity estimation in object detections to align the radial velocity, which is simply back-projecting the radial velocity to the motion direction of the detection. We refer the reader to Fig.~\ref{fig:late_fuse} for an illustration.
It is thus apparent  that the radial velocity is more confident when the angle between the radial direction and motion direction is small, as when it is close to \ang{90}, a very small variance in radial velocity will be exaggerated by back-projection.

Given a set of object detections and \radar\ targets, the key of fully exploiting \radar\ data lies in solving the following two tasks:
(1) {\it association} of  each \radar\ target with the correct object detection for velocity alignment;
(2) {\it aggregation} to combine the velocity estimates from detection and associated \radar\ targets robustly.
Both tasks are non-trivial to solve. The association is not a one-to-one mapping as there are many objects without any associated \radar\ targets, and there are also objects with multiple \radar\ targets. False positives and noisy positions of \radar\ targets also make association difficult. For the aggregation problem, it is hard to estimate the uncertainty of the \radar\ velocity as it also depends on the associated detection.

In this paper, we propose an attention-based mechanism that learns from data to both associate and aggregate. This is illustrated in Fig.~\ref{fig:late_fuse}. Specifically, given pairwise features defined between each object detection and \radar\ target, we first compute pairwise association scores via a learnable matching function. We then aggregate each detection with all \radar\ targets according to the normalized association scores to get the refined velocity estimate. Note that late fusion is performed on {\it dynamic} \radar\ targets only.

\subsubsection{Pairwise Detection-\radar\ Association:}
Given an object detection denoted as $D = (c,x,y,w,l,\theta,\mathbf{v})$ and a \radar\ target denoted as $Q = (\mathbf{q},v_\|,m,t)$, we first define their pairwise feature as follows:
\begin{align}
f(D,Q) &= (f^\text{det}(D), f^\text{det-radar}(D,Q)) \\
\label{eq:feat1}
f^\text{det}(D) &= (w, l, \|\mathbf{v}\|, \frac{v_x}{\|\mathbf{v}\|}, \frac{v_y}{\|\mathbf{v}\|}, \cos(\gamma)) \\
\label{eq:feat2}
f^\text{det-radar}(D,Q) &= (\mathrm{d}x, \mathrm{d}y, \mathrm{d}t, v^\text{bp}) \\
v^\text{bp} &= \min(50, \frac{v_\|}{\cos(\phi)})
\end{align}
where $(\cdot, \cdot)$ indicates  the concatenation operator, $\gamma$ is the angle between $D$'s motion direction and $D$'s radial direction, $\phi$ is the angle between $D$'s motion direction and $Q$'s radial direction, $v^\text{bp}$ is the back-projected radial velocity (capped by 50 m/s to avoid very large values), and $(\mathrm{d}x, \mathrm{d}y, \mathrm{d}t)$ are the offsets in BEV positions and timestamps of $D$ and $Q$.

We then compute the pairwise association score by feeding the above feature to a learnable matching function:
\begin{equation}
s_{i,j} = \text{MLP}_\text{match}(f(D_i, Q_j))
\end{equation}
In our case the matching function is parameterized as a Multi-Layer Perceptron (MLP) with five layers  with  32, 64, 64, 64 and 1 channels respectively.

\subsubsection{Velocity Aggregation:}
We compute the association scores for all detections and \radar\ target pairs and refine the velocity estimate of each detection $D_i$ by aggregating information from all \radar\ targets.
Towards this goal, we first normalize the association scores of all Radar targets to sum to 1. We append an additional score of $1$ before normalization to handle cases with no association.
\begin{equation}
\mathbf{s}^\text{norm}_i = \texttt{softmax}((1, s_{i,:}))
\end{equation}
We then refine the velocity magnitude by summing all the candidates (the detection itself as well as all \radar\ targets) weighted by their normalized scores:
\begin{equation}
v_i' = \mathbf{s}^\text{norm}_i \cdot (\|\mathbf{v}_i\|, v^\text{bp}_{i,:})^\top
\end{equation}
The 2D velocity estimate is then computed as the refined velocity magnitude:
\begin{equation}
\mathbf{v}'  = v' \cdot (\frac{v_x}{\|\mathbf{v}\|}, \frac{v_y}{\|\mathbf{v}\|})
\end{equation}
where the detection index $i$ is omitted for brevity.

\subsection{Learning and Inference}
We trained the proposed \lidar\ and \radar\ fusion model with a multi-task loss defined as a weighted sum of the detection loss, velocity loss on the detection output, as well as the  velocity loss on the late fusion output:
\begin{equation}
\mathcal{L} = (\mathcal{L}^\text{det}_\text{cls} + \alpha\cdot\mathcal{L}^\text{det}_\text{reg}) + \beta\cdot(\mathcal{L}^\text{velo}_\text{cls} + \mathcal{L}^\text{velo}_\text{reg}) + \delta\cdot\mathcal{L}^{\text{velo}\_\text{attn}}_\text{reg}
\end{equation}
where $\mathcal{L}^\text{det}_\text{cls}$ is the cross-entropy loss on classification score $c$, $\mathcal{L}^\text{det}_\text{reg}$ is  the smooth $\ell_1$ loss summed over the position, size and orientation terms, $\mathcal{L}^\text{velo}_\text{cls}$ is the cross-entropy loss on moving probability $m$, $\mathcal{L}^\text{velo}_\text{reg}$ is the smooth $\ell_1$ loss on $\mathbf{v}$, and $\mathcal{L}^{\text{velo}\_\text{attn}}_\text{reg}$ is the smooth $\ell_1$ loss on $\mathbf{v}'$. $\alpha$, $\beta$ and $\delta$ are scalars that balance different tasks.
Note that we do not require explicit supervision to learn object and \radar\ association, which is an advantage of the attention-based late fusion module where the association is implicitly learned.

We use the Adam optimizer \cite{adam} with batch normalization \cite{ioffe2015batch} after every convolution layer and layer normalization \cite{ba2016layer} after every fully-connected layer (except for the final output layer). For detection we use hard negative mining. $\mathcal{L}^\text{det}_\text{reg}$, $\mathcal{L}^\text{velo}_\text{cls}$ and $\mathcal{L}^\text{velo}_\text{reg}$ are computed on positive samples only, and  $\mathcal{L}^{\text{velo}\_\text{attn}}_\text{reg}$ is computed on true positive detections only.
We apply the same post-processing to generate final detections during training and testing phases, where the top 200 detections per class are kept and NMS is applied thereafter.

% !TEX root = top.tex
\section{Experimental Evaluation}
\subsection{Datasets and Evaluation Metrics}
\subsubsection{nuScenes:}
We validate the proposed method on the nuScenes dataset \cite{caesar2019nuscenes}. This dataset contains sensor data from 1 \lidar\ and 5 \radar s, with object labels at 2Hz. Velocity labels are computed as finite difference between consecutive frames. Since we focus on dynamic objects, we evaluate on two challenging object classes: cars and motorcycles, as their velocities have high variance.
We follow the official training/validation split with 700/150 logs each.
We report the model performance on object detection and velocity estimation. Average Precision (AP) is used as the detection metric, which is defined on center distance in BEV between the detection and the label. The final AP is averaged over four different distance thresholds (0.5m, 1m, 2m and 4m). Average Velocity Error (AVE) is used as the velocity metric, which is computed as the $\ell_2$ velocity error averaged over all true positive detections (at 2m threshold). Cars are evaluated within 50m range, while motorcycles are evaluated within 40m range. Labels with 0 \lidar\ and \radar\ points are ignored.

\subsubsection{DenseRadar:}
One advantage of \radar\ over \lidar\ is its longer sensing range. To showcase this, we further evaluate our model on a self-collected dataset, called {\it DenseRadar}, with vehicle labels within 100m range for 5002 snippets. Velocity labels are estimated by fitting a kinematic bicycle model to the trajectory, which produces smoother velocities compared with the finite difference procedure employed in nuScenes. We use similar metrics as nuScenes. For detection we compute AP at 0.7 IoU in BEV. For velocity we report Average Dynamic Velocity Error (ADVE)  on {\it dynamic} objects only. We make a training/validation split with 4666/336 logs each.

\subsection{Implementation Details}
We train a two-class model on nuScenes with a shared backbone network and class-specific detection headers. Global data augmentation is used during training, with random translations from [-1, 1]m in the X and Y axes and [-0.2, 0.2]m in the Z axis, random scaling from [0.95, 1.05], random rotation from [-\ang{45}, \ang{45}] along the Z axis, and random left-right and front-back flipping. We do not apply augmentation at test time. To alleviate the class imbalance, we duplicate training frames that contain motorcycles by 5 times. The model is trained for 25 epochs with a batch size of 32 frames on 8 GPUs.
We use an input voxel size of 0.125m in the X and Y axes, and 0.2m in the Z axis. We use $\alpha=1$ and $\beta=\delta=0.1$. Hyper-parameter tuning is conducted on the train-detect/train-track split.

We train a single-class model on DenseRadar. Since the dataset is much larger, we do not apply data augmentation. We use an input voxel resolution of 0.2m in all three axes due to the extra computation due to the longer detection range. We use $\alpha=1$ and $\beta=\delta=0.5$. The model is trained for 1.5 epochs.

\subsection{Comparison with the State-of-the-Art}
% !TEX root = ../top.tex

\setlength{\tabcolsep}{6pt}
\begin{table}[t]
\begin{center}
\caption{{\bf Comparison with the state-of-the-art on nuScenes validation set}}
\label{tab:main_results}
\begin{tabular}{l|c|ccc|ccc}
\hline
\multirow{2}{*}{Method} & \multirow{2}{*}{Input} & \multicolumn{3}{c|}{Cars} & \multicolumn{3}{c}{Motorcycles} \\
 & & AP$\uparrow$ & AP@2m$\uparrow$ & AVE$\downarrow$ & AP$\uparrow$ & AP@2m$\uparrow$ & AVE$\downarrow$ \\
\hline
MonoDIS \cite{monodis} & I & 47.8 & 64.9 & - & 28.1 & 37.7 & - \\
PointPillar \cite{lang2019pointpillars} & L & 70.5 & 76.1 & 0.269 & 20.0 & 22.8 & 0.603 \\
PointPillar+ \cite{vora2020pointpainting} & L & 76.7 & 80.5 & 0.209 & 35.0 & 38.6 & 0.371 \\
PointPainting \cite{vora2020pointpainting} & L+I & 78.8 & 82.9 & 0.206 & 44.4 & 48.1 & 0.351 \\
3DSSD \cite{yang20203dssd} & L & 81.2 & 85.8 & 0.188 & 36.0 & 39.9 & 0.356 \\
CBGS \cite{zhu2019class} & L & 82.3 & 85.9 & 0.230 & 50.6 & 52.4 & 0.339 \\
\hline
RadarNet (Ours) & L+R & {\bf 84.5} & {\bf 87.9} & {\bf 0.175} & {\bf 52.9} & {\bf 55.6} & {\bf 0.269} \\
\hline
\end{tabular}
\end{center}
\end{table}

We compare our \lidar\ and \radar\ fusion model with other state-of-the-art perception models on nuScenes and show the evaluation results in Table \ref{tab:main_results}. Specifically, we compare with the camera-based method MonoDIS \cite{monodis}, the \lidar-based methods PointPillar \cite{lang2019pointpillars}, PointPillar+ \cite{vora2020pointpainting}, 3DSSD \cite{yang20203dssd}, CBGS \cite{zhu2019class}, and the \lidar\ and camera fusion method PointPainting \cite{vora2020pointpainting}.
RadarNet outperforms all methods significantly in both detection AP and velocity error. Compared with the second best on cars/motorcycles, our model shows an absolute gain of 2.2\%/2.3\% in detection AP and a relative reduction of 7\%/21\% in velocity error.

\subsection{Ablation Study}
We conduct an ablation study on the nuScenes and DenseRadar datasets to validate the effectiveness of our two-level fusion scheme.
To better verify the advantage of the proposed attention-based late fusion, we build a strong baseline with carefully designed heuristics.
Recall that our attention-based late fusion consists of two steps: association and aggregation. As a counterpart, we build the baseline fusion method by replacing each step with heuristics. In particular, for each detection candidate, we first use a set of rules to determine the \radar\ targets associated with it. Given a set of associated \radar\ targets (if any), we then take the median of their aligned velocities (by back-projecting to the motion direction of the detection) as the estimate from \radar\ and average it with the initial velocity estimate of the detection. If there are no associated \radar\ targets, we keep the original detection velocity.

Below we define the set of rules we designed for determining the associated \radar\ targets. Given the features in Eq. \ref{eq:feat1} and Eq. \ref{eq:feat2}, a \radar\ target is considered as associated if it meets all of the following conditions:
\begin{align}
\sqrt{(\mathrm{d}x)^2 + (\mathrm{d}y)^2} &< 3 \text{ m} \\
\gamma &< \ang{40} \\
\|\mathbf{v}\| &> 1 \text{ m/s}\\
v^\text{bp} &< 30 \text{ m/s}
\end{align}
We define these rules to filter out unreliable \radar\ targets, and the thresholds are chosen via cross-validation.

% !TEX root = ../top.tex

% \setlength{\tabcolsep}{2pt}
\begin{table}[t]
\begin{center}
\caption{{\bf Ablation study on nuScenes validation set}}
\label{tab:ab_nu}
\begin{tabular}{l|ccc|cc|cc}
\hline
\multirow{2}{*}{Model} & \multirow{2}{*}{LiDAR} & \multicolumn{2}{c|}{Radar} & \multicolumn{2}{c|}{Cars} & \multicolumn{2}{c}{Motorcycles} \\
 & & Early & Late & AP@2m$\uparrow$ & AVE$\downarrow$ & AP@2m$\uparrow$ & AVE$\downarrow$ \\
\hline
LiDAR & \checkmark & - & - & 87.6 & 0.203 & 53.7 & 0.316 \\
\hline
Early & \checkmark & \checkmark & - & +0.3 & -2\% & +1.9 & -0\% \\
Heuristic & \checkmark & \checkmark & heuristic & +0.3 & -9\% & +1.9 & -4\% \\
RadarNet & \checkmark & \checkmark & attention & {\bf +0.3} & {\bf -14\%} & {\bf +1.9} & {\bf -15\%} \\
\hline
\end{tabular}
\end{center}
\end{table}

% !TEX root = ../top.tex

% \setlength{\tabcolsep}{4pt}
\begin{table}[t]
\begin{center}
\caption{{\bf Ablation study on DenseRadar validation set}}
\label{tab:ab_xe}
\begin{tabular}{l|ccc|ccc|c}
\hline
\multirow{2}{*}{Model} & \multirow{2}{*}{LiDAR} & \multicolumn{2}{c|}{Radar} & \multicolumn{3}{c|}{Vehicles AP $\uparrow$} & \multirow{2}{*}{ADVE $\downarrow$}\\
 & & Early & Late & 0-40m & 40-70m & 70-100m & \\
\hline
LiDAR & \checkmark & - & - & 95.4 & 88.0 & 77.5 & 0.285 \\
\hline
Early & \checkmark & \checkmark & - & +0.3 & +0.5 & +0.8 & -3\% \\
Heuristic & \checkmark & \checkmark & heuristic & +0.3 & +0.5 & +0.8 & -6\% \\
RadarNet & \checkmark & \checkmark & attention & {\bf +0.3} & {\bf +0.5} & {\bf +0.8} & {\bf -19\%} \\
\hline
\end{tabular}
\end{center}
\end{table}

\subsubsection{Evaluation on nuScenes:}
We show ablation results on nuScenes in Table \ref{tab:ab_nu}. Note that our \lidar\ only model  already achieves state-of-the-art performance.
Adding early fusion improves detection of motorcycles by 1.9\% absolute AP, as the \lidar\ observations are sparse and therefore \radar\ data serves as  additional evidence. Early fusion does not affect the velocity performance much as only density information is exploited at present.
When it comes to late fusion, our approach achieves over 14\% velocity error reduction, significantly outperforming the heuristic baseline especially in motorcycles, where we typically have few \radar\ targets and therefore more noise.

\begin{figure}[t]
\centering
\includegraphics[width=0.5\linewidth]{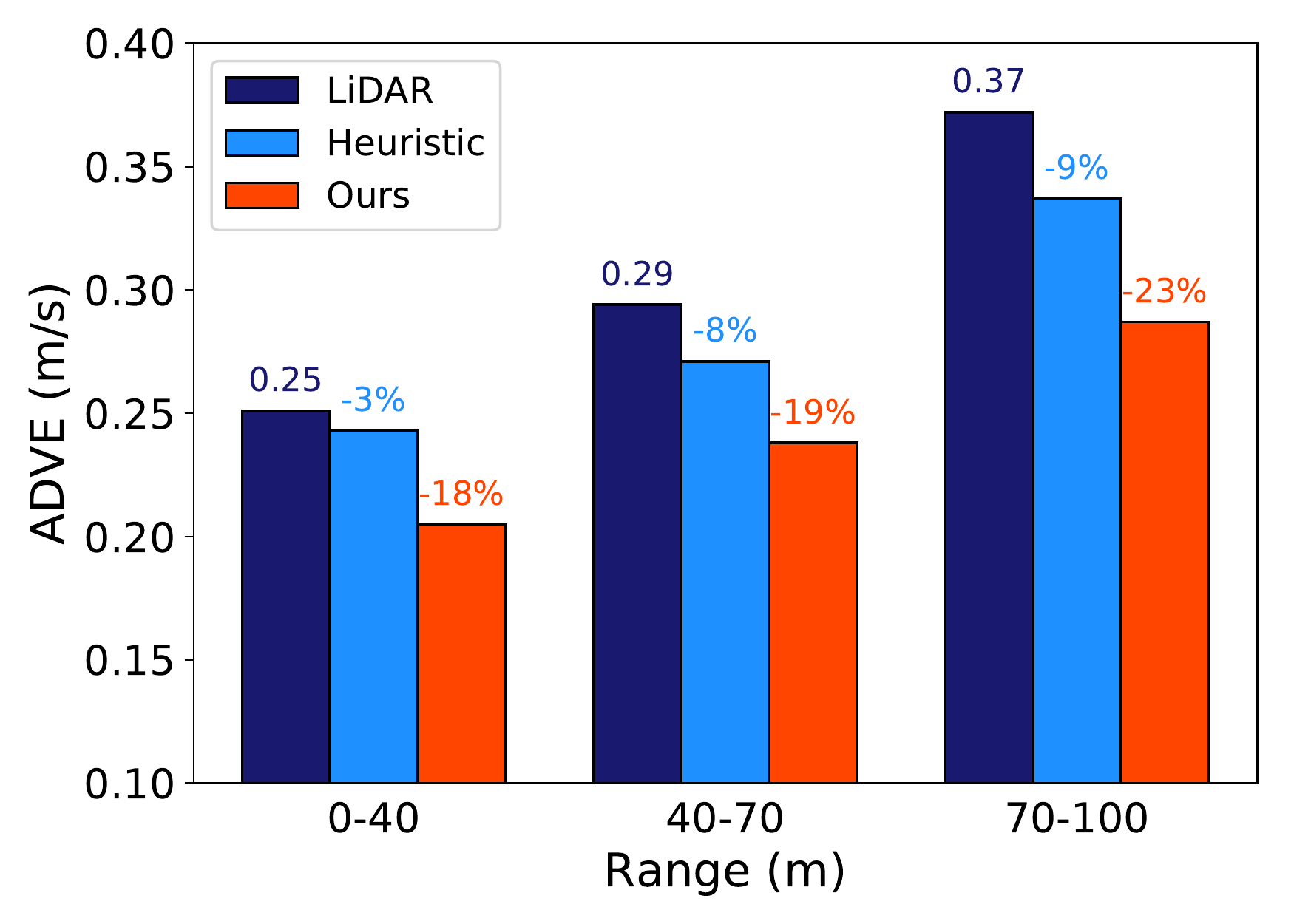}\includegraphics[width=0.5\linewidth]{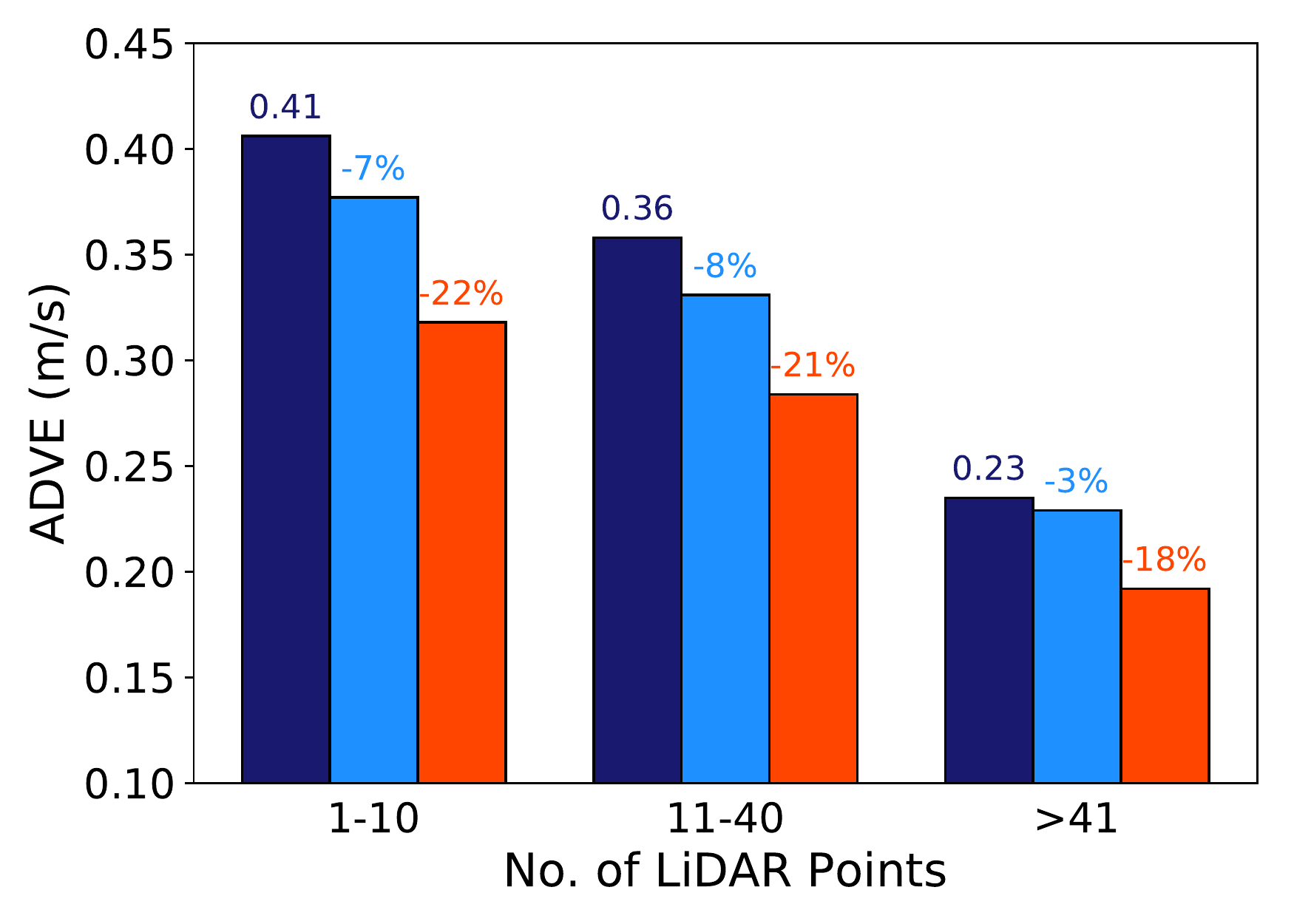}
\includegraphics[width=0.5\linewidth]{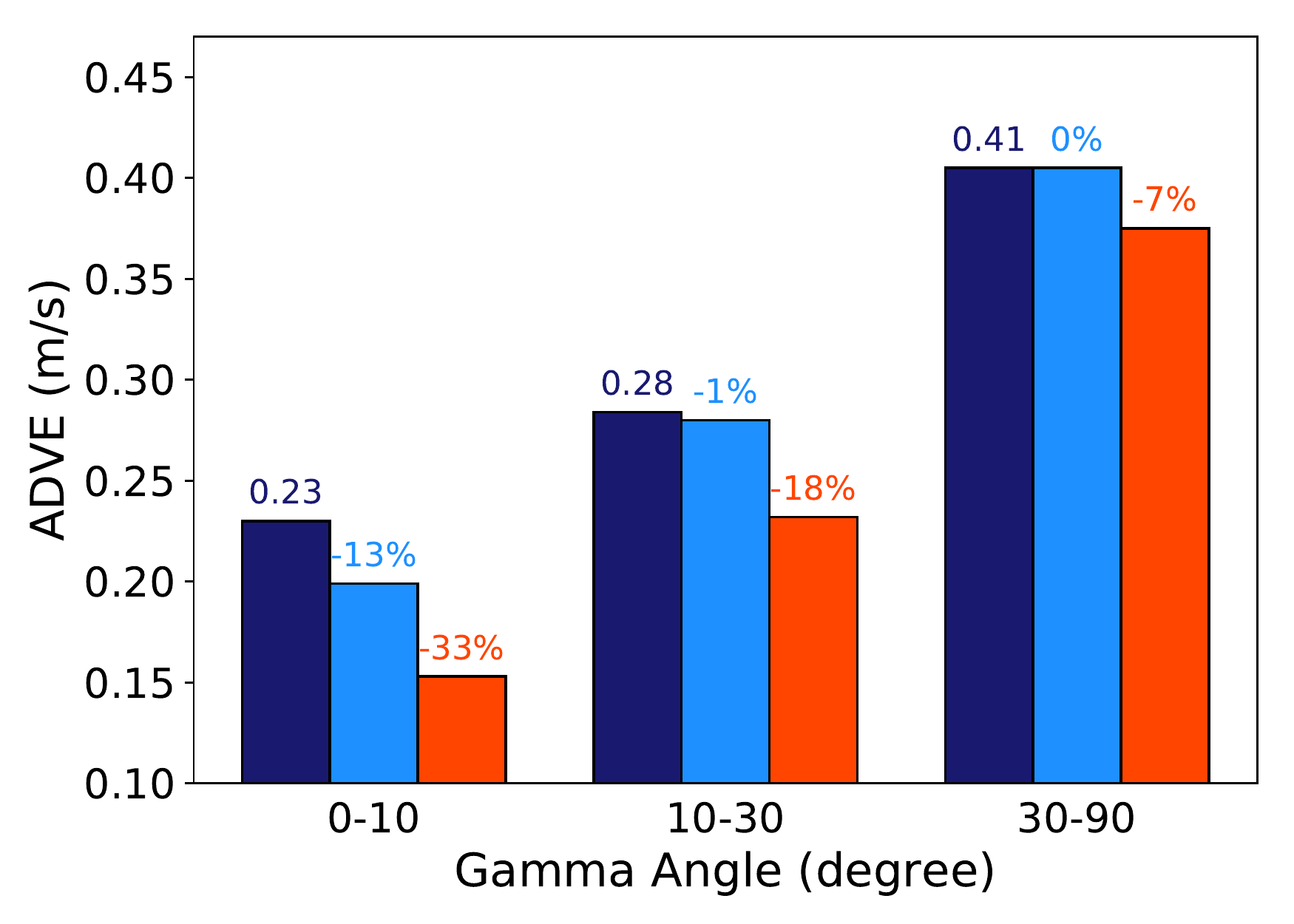}\includegraphics[width=0.5\linewidth]{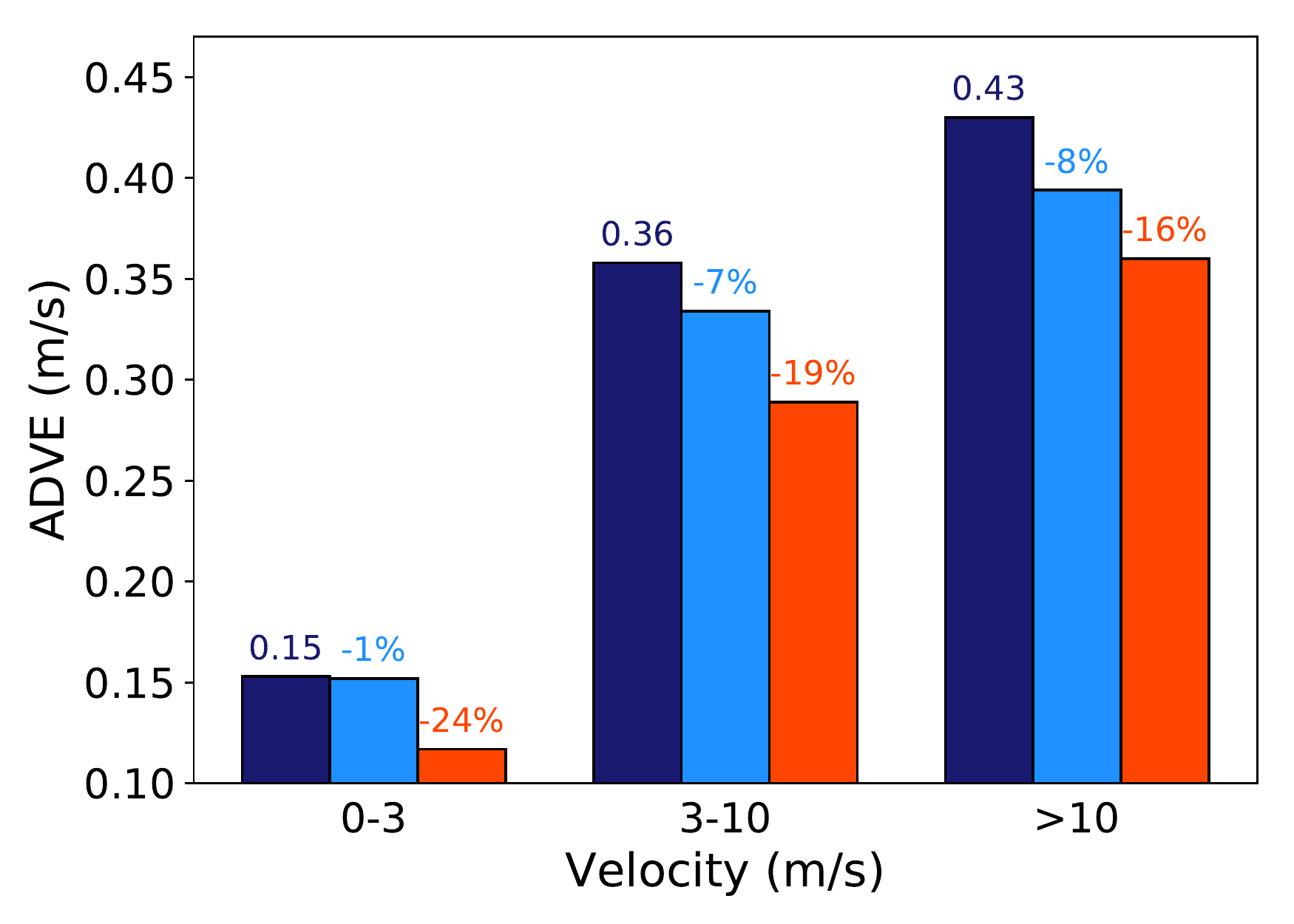}
\caption{{\bf Fine-grained evaluation of velocity estimation on DenseRadar validation set.}}
\label{fig:ablation}
\end{figure}

\subsubsection{Evaluation on DenseRadar:}
Ablation results on DenseRadar are depicted in Table \ref{tab:ab_xe}. We show detection APs in near range (0-40m), mid range (40-70m) and long range (70-100m) respectively.
Early fusion helps long-distance object detection, bringing 0.8\% absolute gain in the 70-100m range detection AP.
When late fusion is added, larger improvements are achieved than on nuScenes (from 14\% to 19\%). Two reasons may account for this: (1) DenseRadar uses higher-end \radar\ sensors that produce denser returns; (2) we evaluate in longer range (100m vs. 50m), which is more challenging and therefore there is more room for improvement. However, the heuristic baseline still gets lower than 10\% gain, showing the advantage of the proposed attention-based mechanism which can learn from noisy data.

\subsection{Fine-Grained Analysis}

\begin{figure}[t]
\centering
\includegraphics[width=0.24\linewidth,trim={50cm 30cm 8cm 30cm},clip]{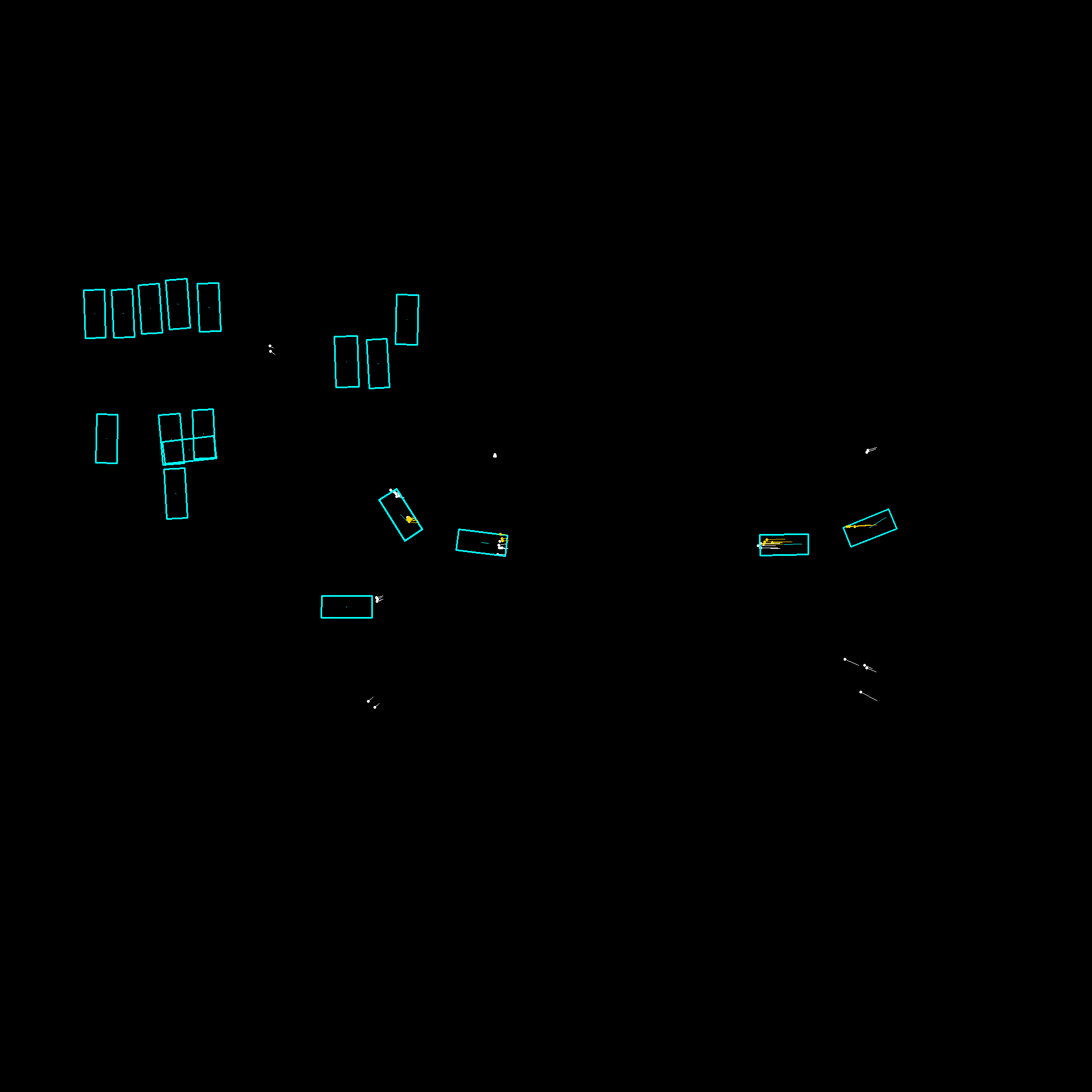}
\includegraphics[width=0.24\linewidth,trim={52cm 31cm 6cm 29cm},clip]{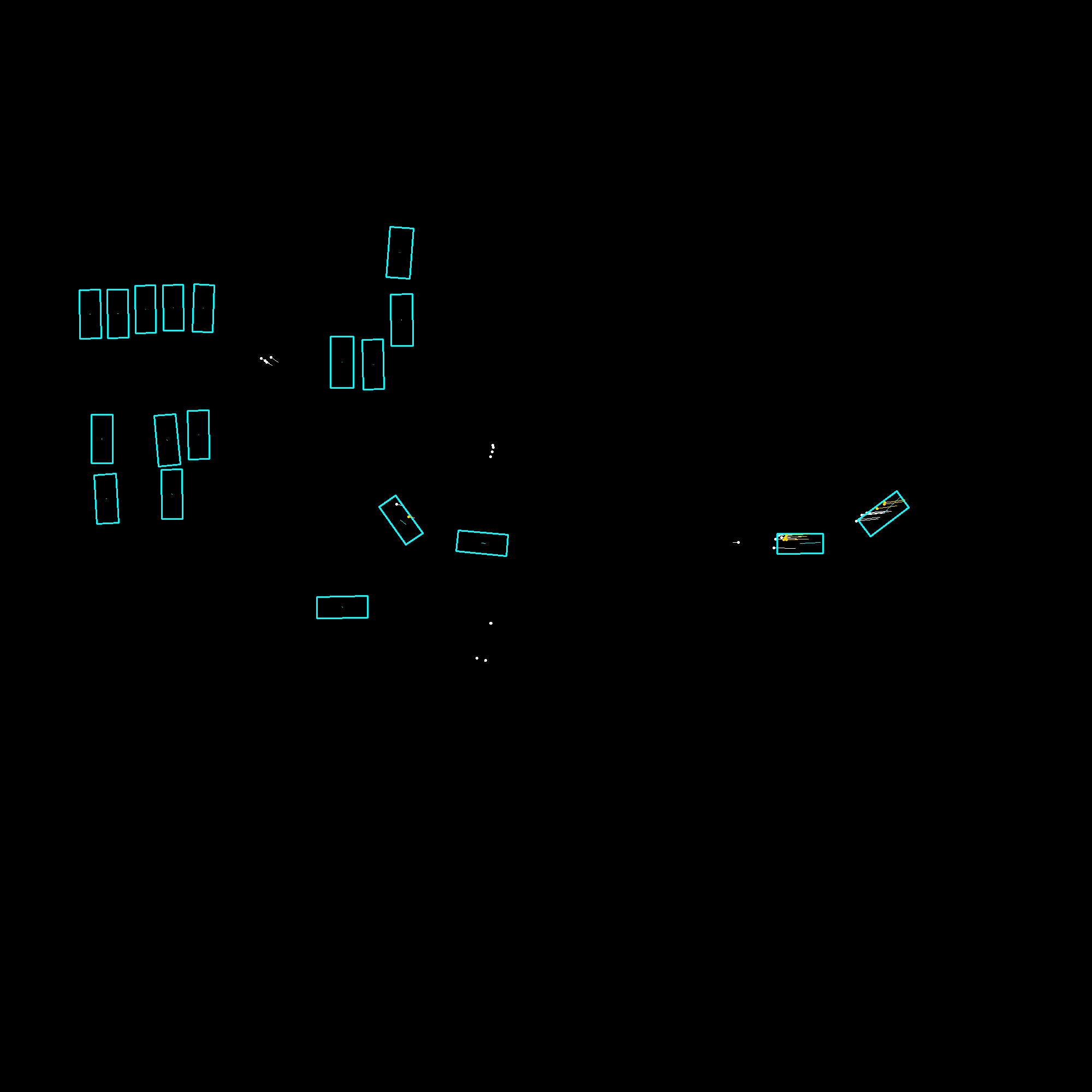}
\includegraphics[width=0.24\linewidth,trim={53cm 32cm 5cm 28cm},clip]{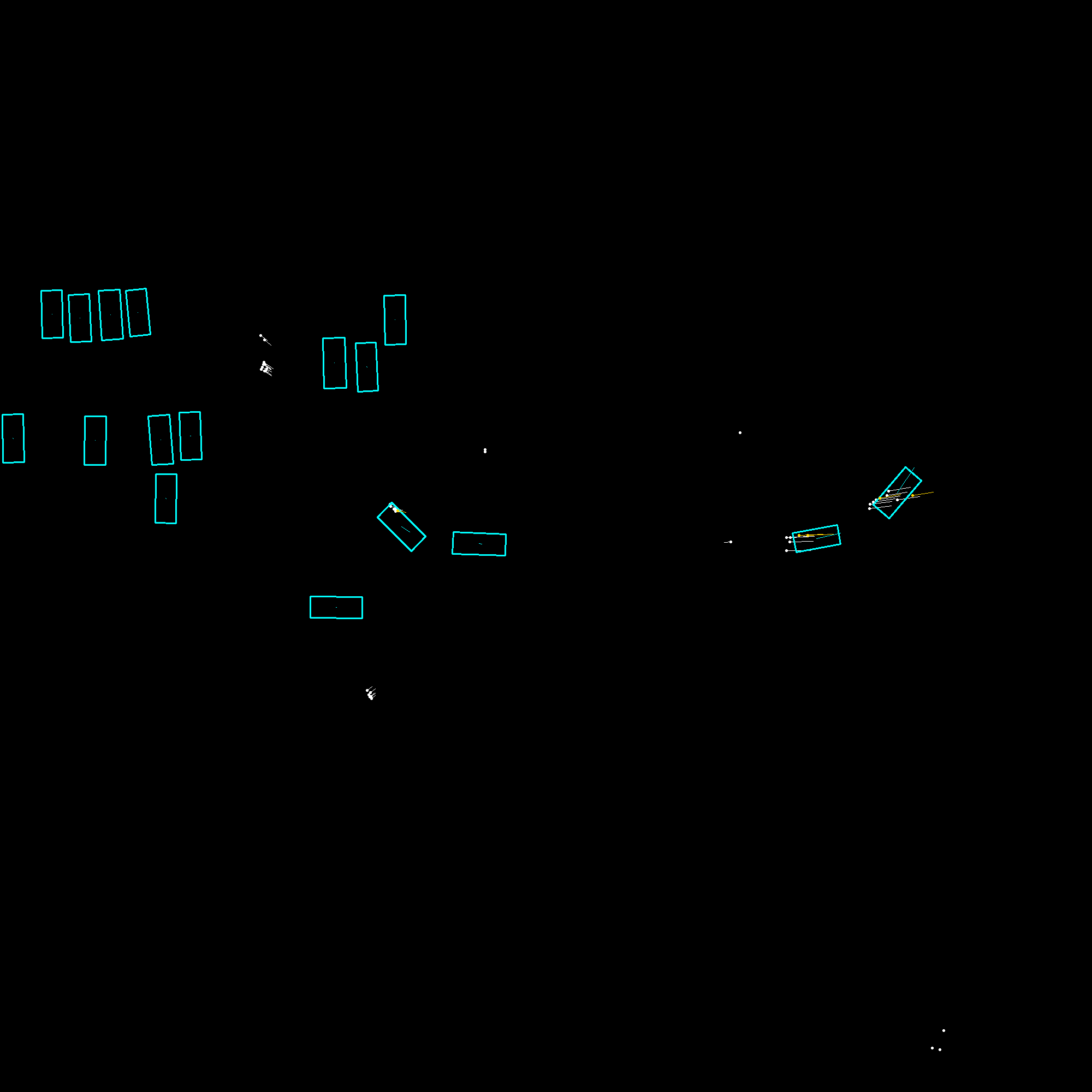}
\includegraphics[width=0.24\linewidth,trim={53cm 35cm 5cm 25cm},clip]{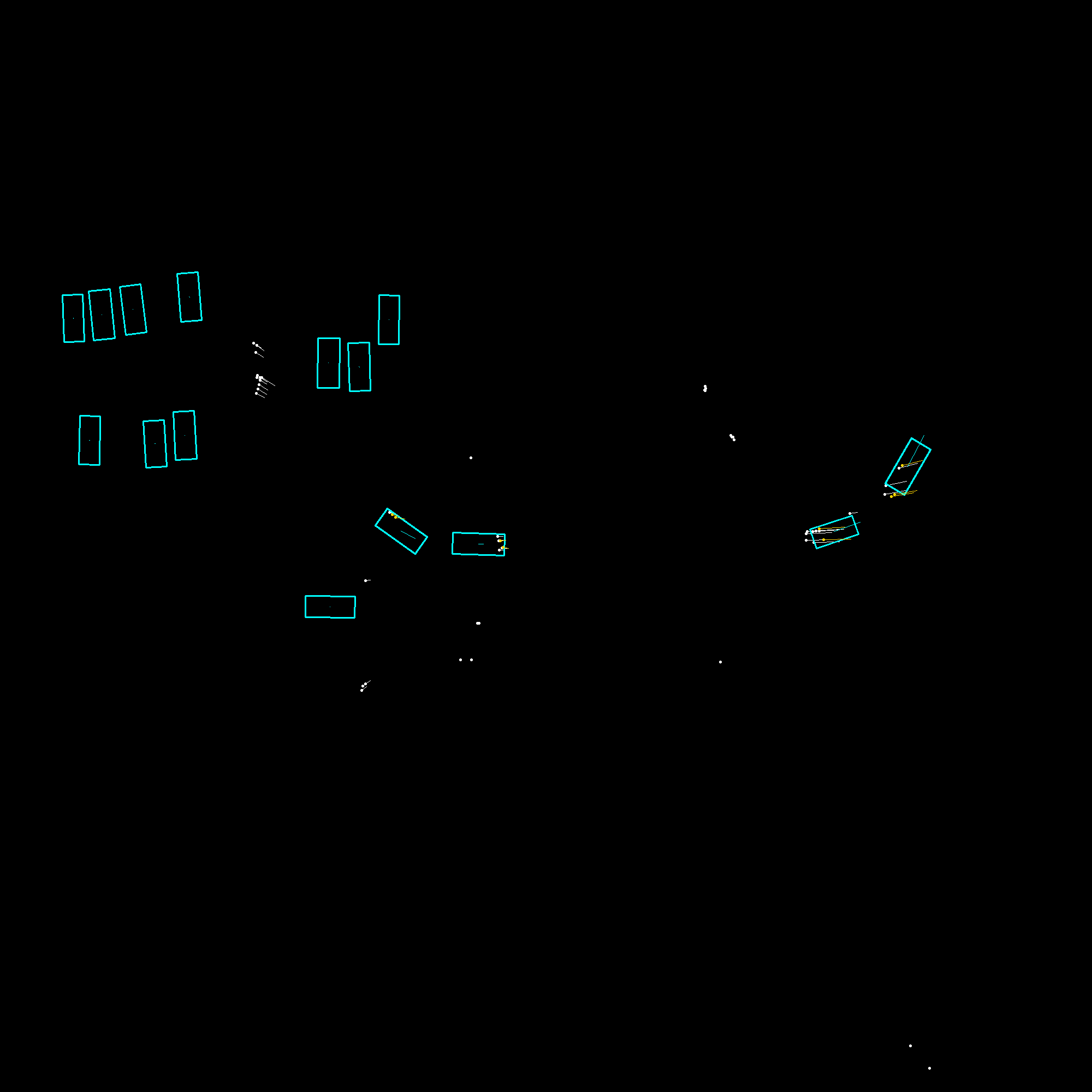}
\includegraphics[width=0.24\linewidth,trim={20cm 36cm 38cm 24cm},clip]{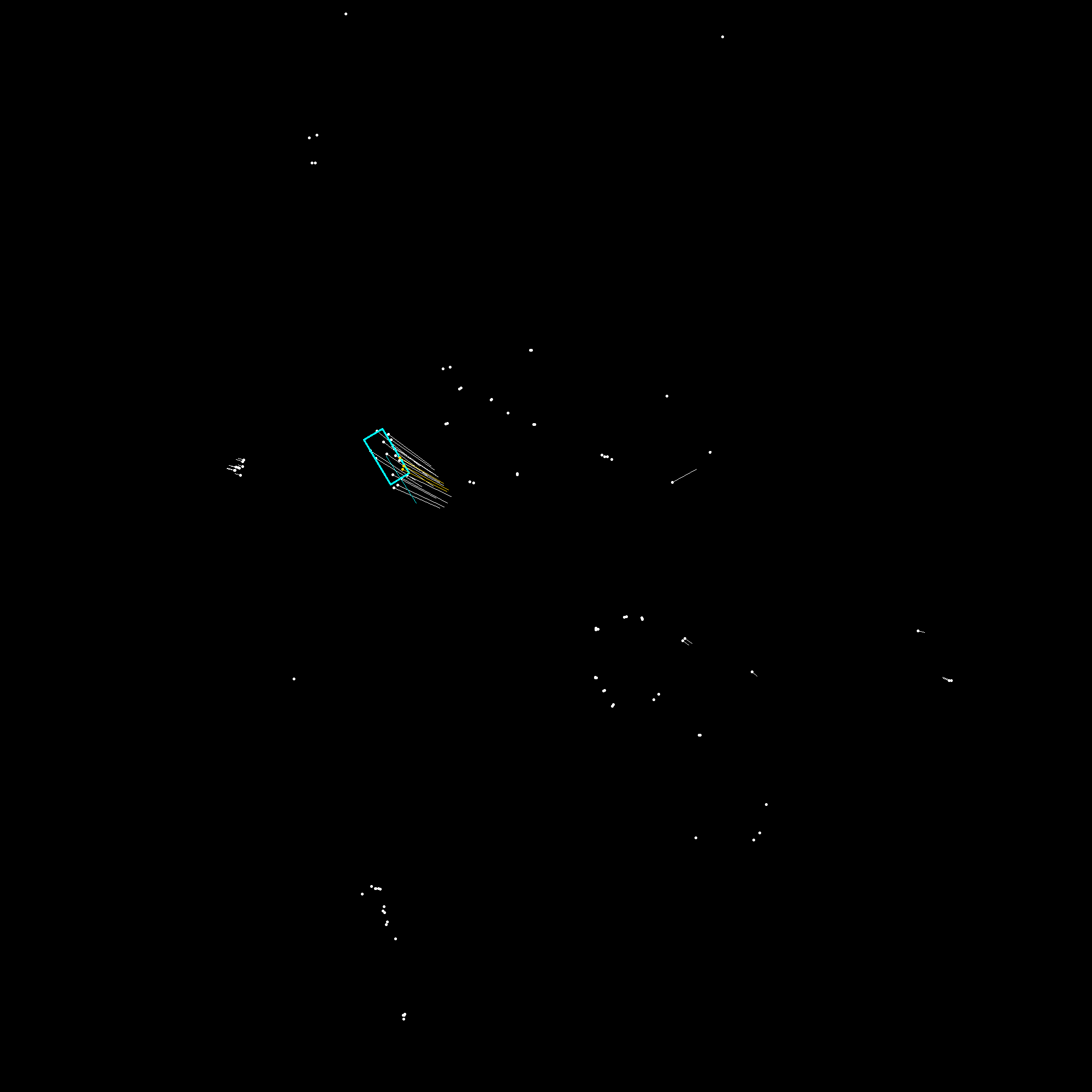}
\includegraphics[width=0.24\linewidth,trim={20cm 33cm 38cm 27cm},clip]{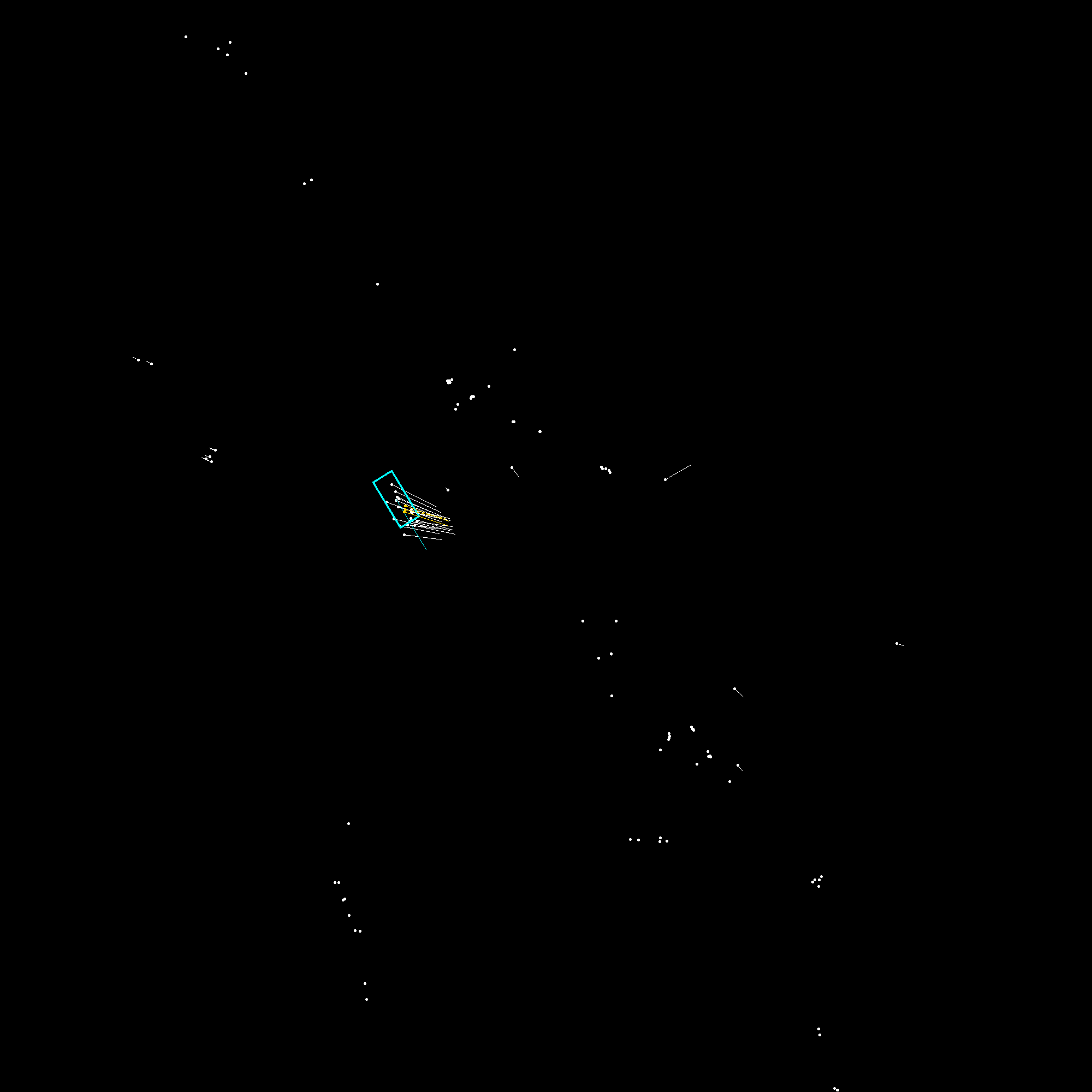}
\includegraphics[width=0.24\linewidth,trim={20cm 30cm 38cm 30cm},clip]{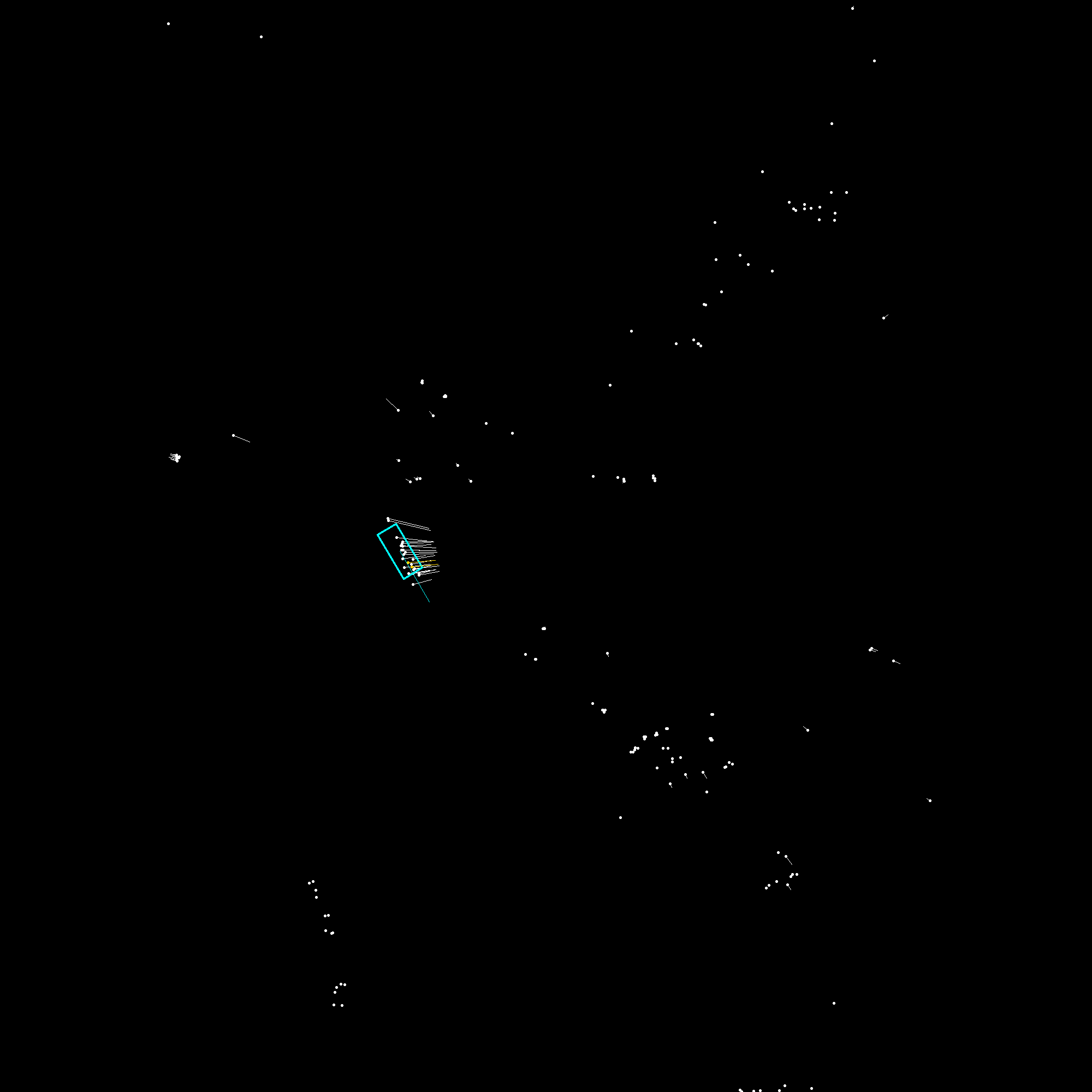}
\includegraphics[width=0.24\linewidth,trim={20cm 27cm 38cm 33cm},clip]{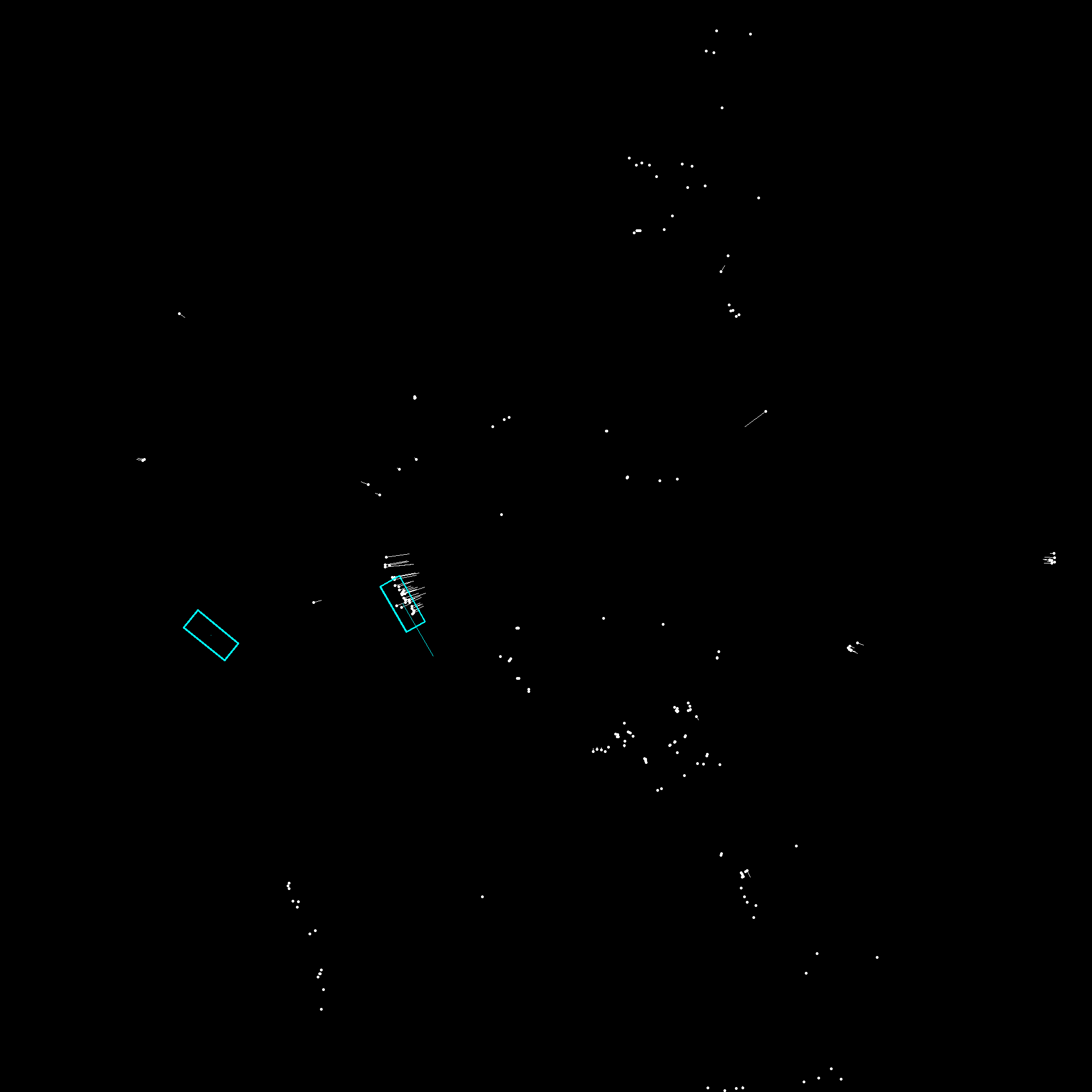}
\includegraphics[width=0.24\linewidth,trim={46cm 25cm 12cm 35cm},clip]{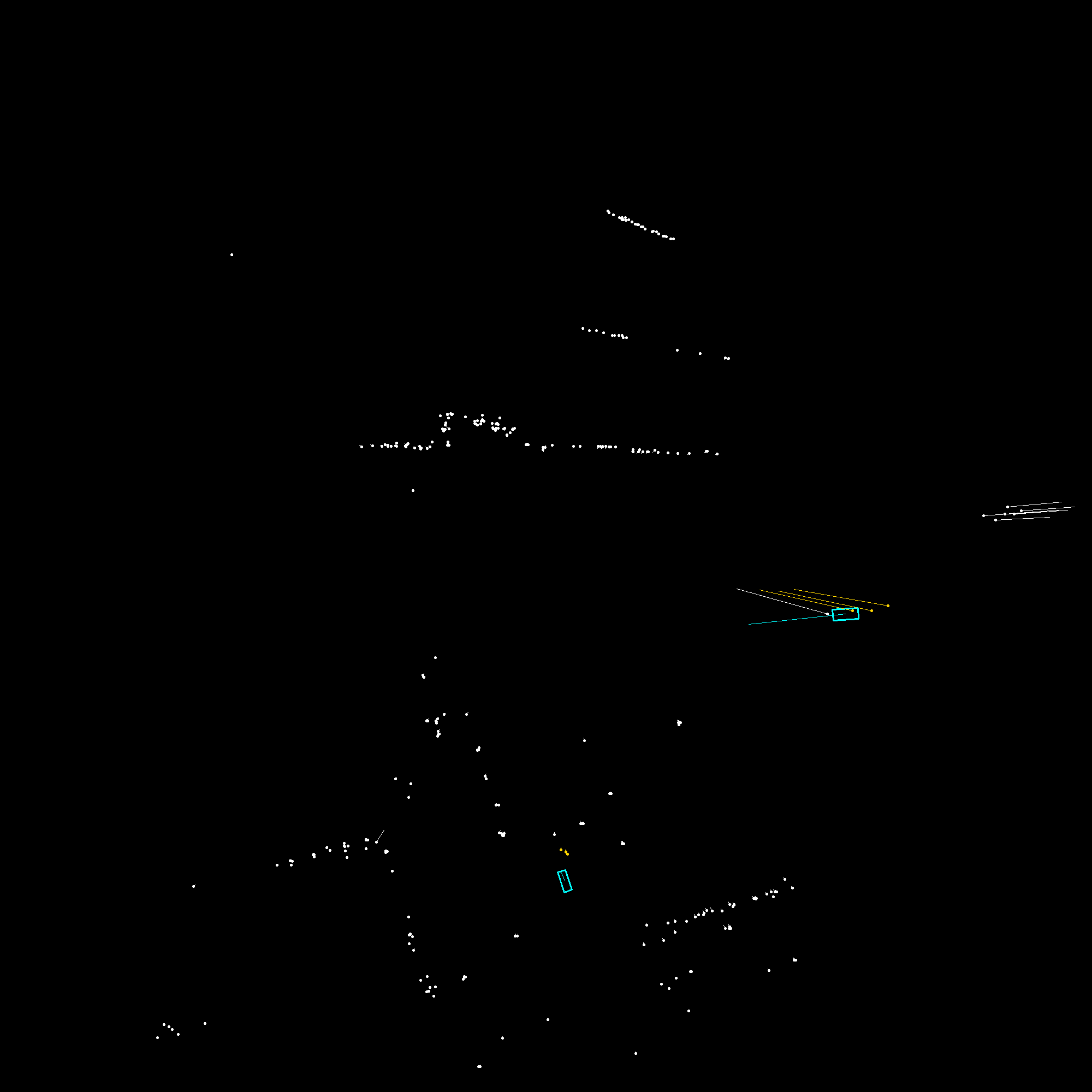}
\includegraphics[width=0.24\linewidth,trim={36cm 25cm 22cm 35cm},clip]{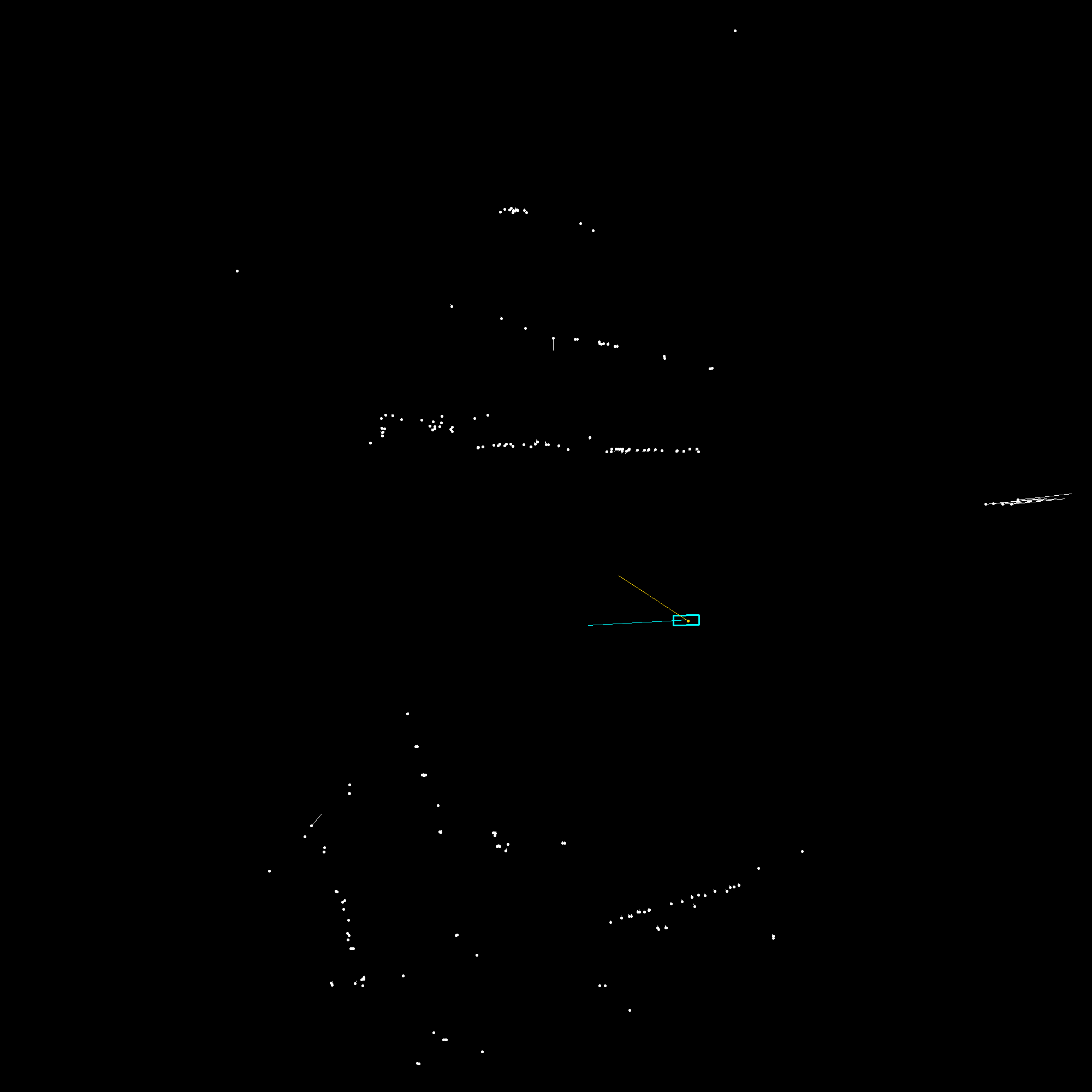}
\includegraphics[width=0.24\linewidth,trim={26cm 25cm 32cm 35cm},clip]{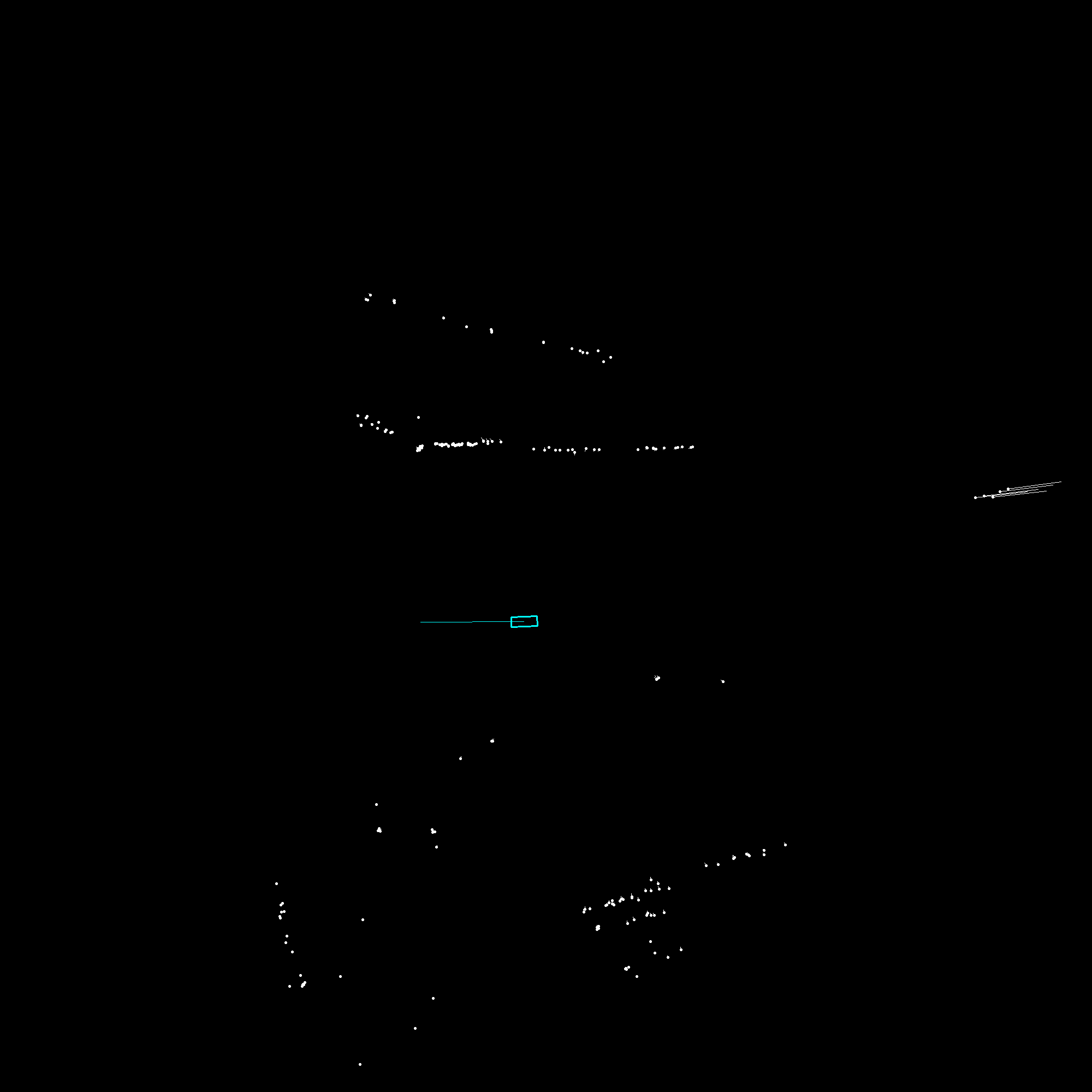}
\includegraphics[width=0.24\linewidth,trim={6cm 25cm 52cm 35cm},clip]{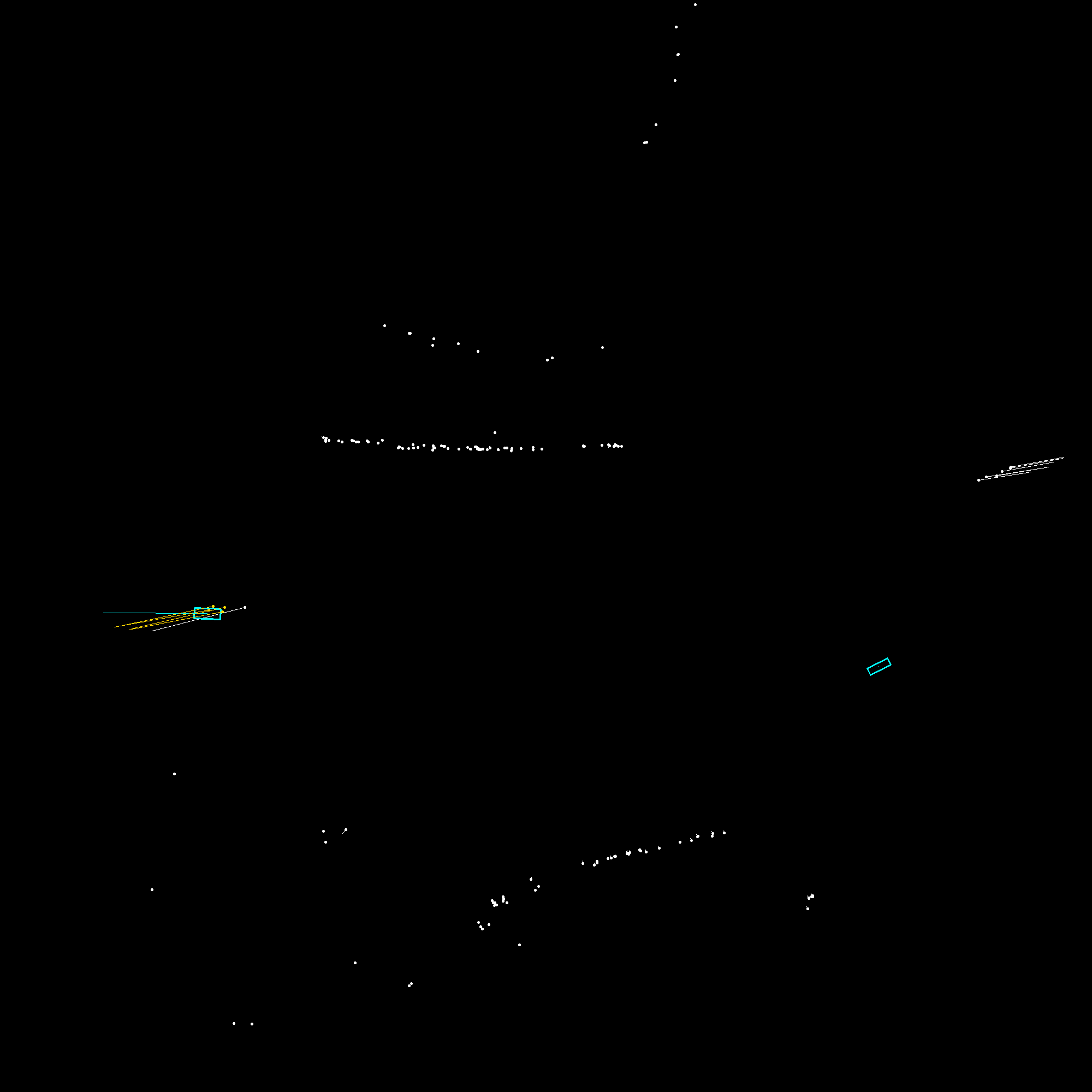}
\includegraphics[width=0.24\linewidth,trim={44cm 34cm 14cm 26cm},clip]{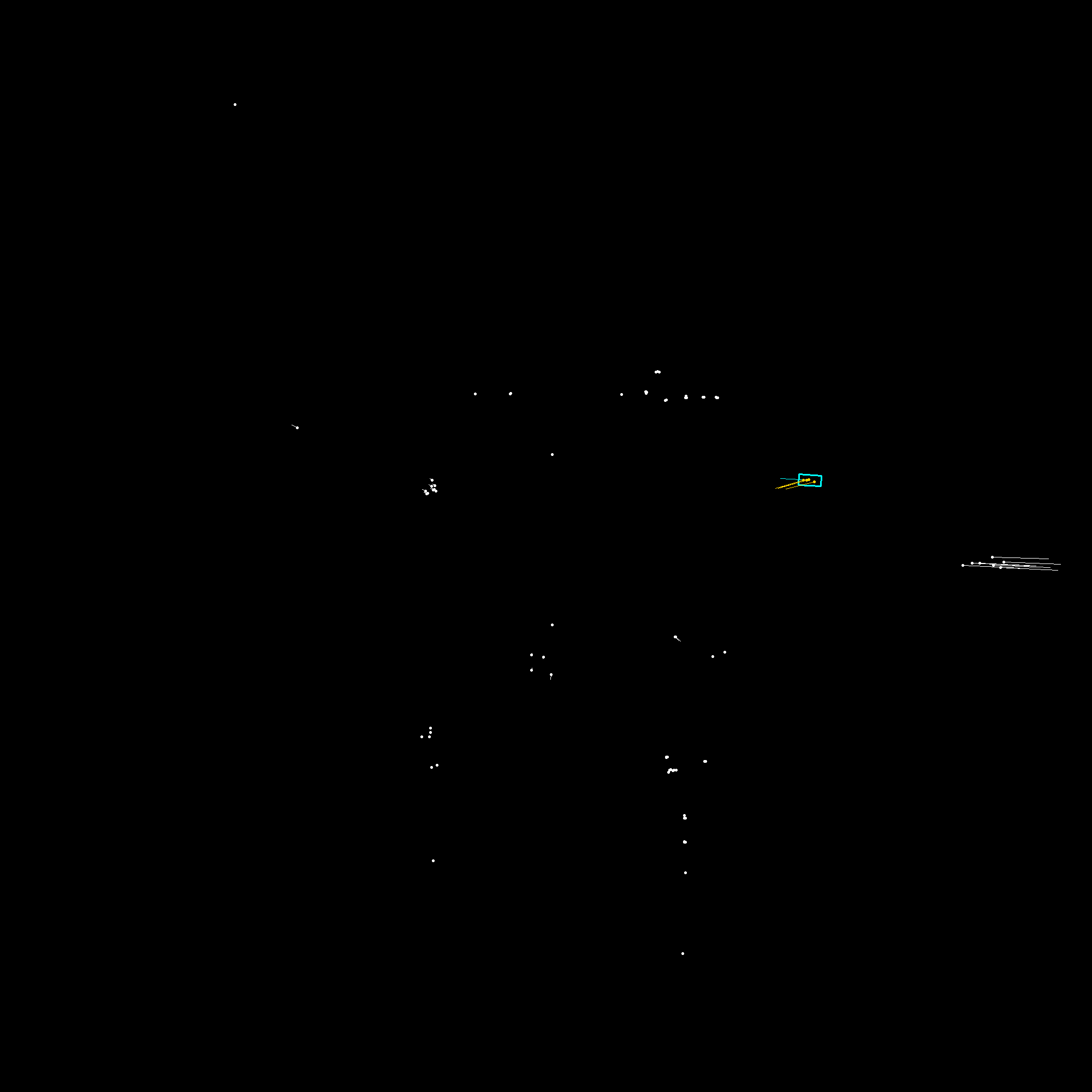}
\includegraphics[width=0.24\linewidth,trim={40cm 34cm 18cm 26cm},clip]{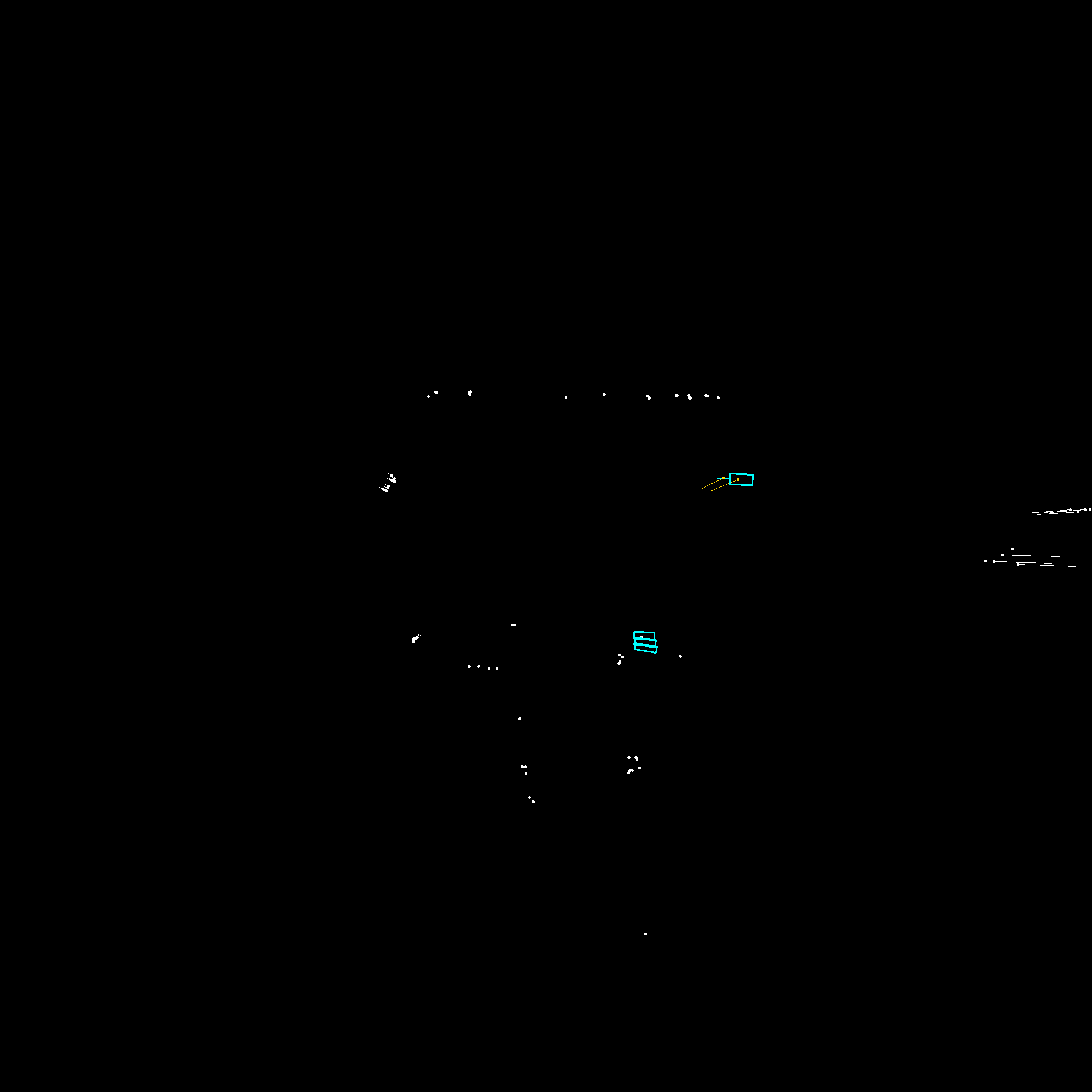}
\includegraphics[width=0.24\linewidth,trim={37cm 34cm 21cm 26cm},clip]{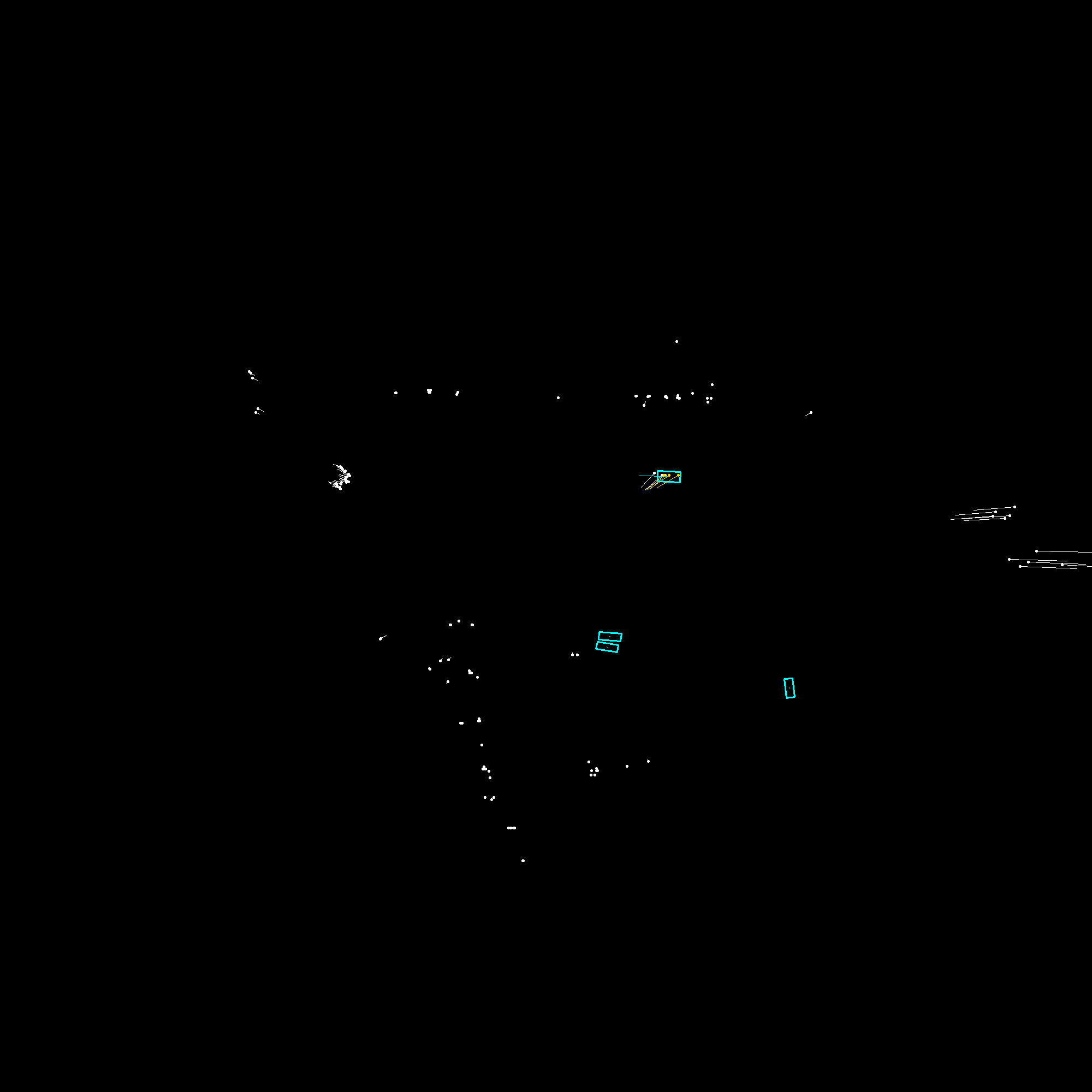}
\includegraphics[width=0.24\linewidth,trim={20cm 34cm 38cm 26cm},clip]{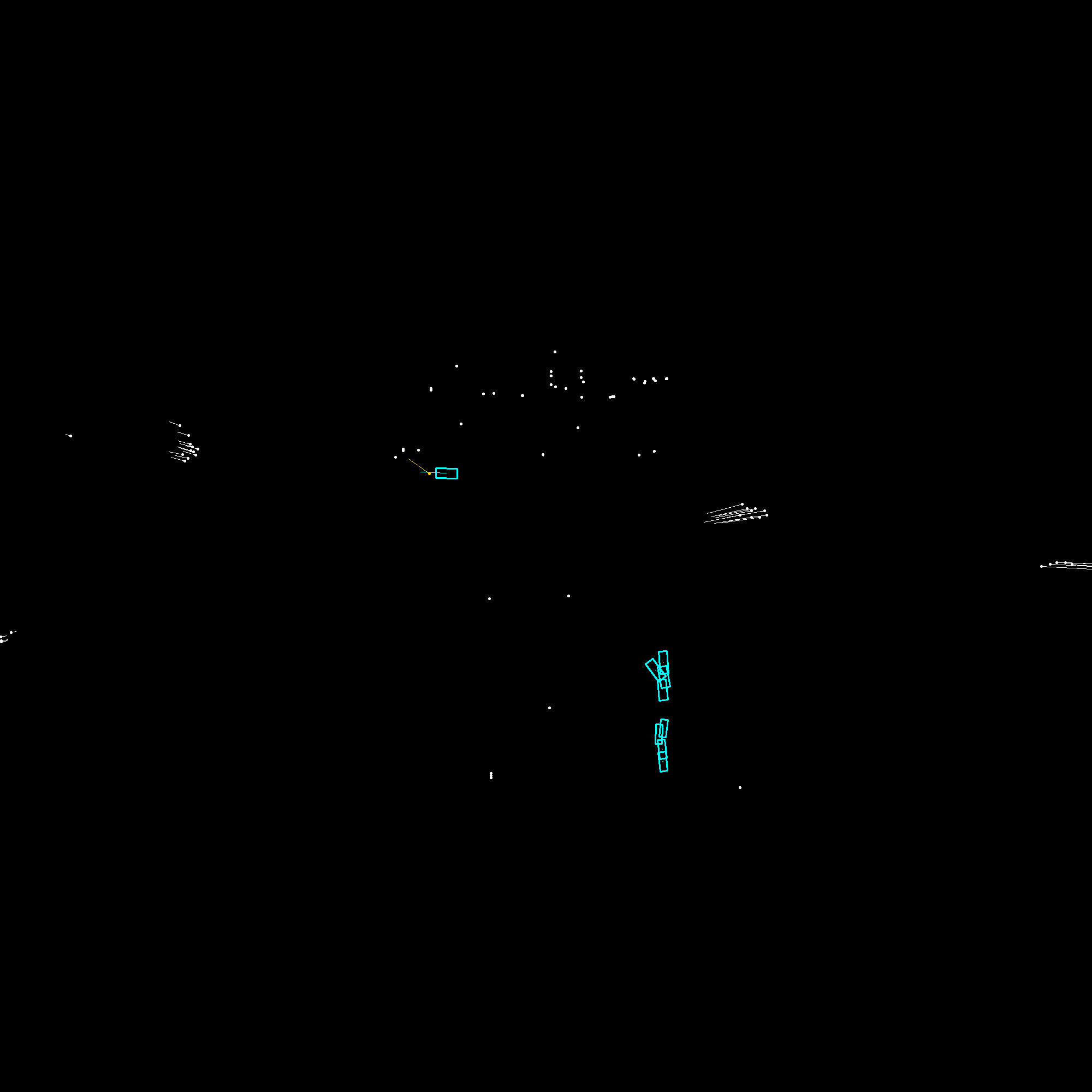}
\fbox{times step = 1} \fbox{ } \fbox{times step = 2} \fbox{ } \fbox{times step = 3} \fbox{ } \fbox{times step = 4}
\caption{{\bf Qualitative Results:} Visualization of learned detections and \radar\ associations for cars (row 1 \& 2) and motorcycles (row 3 \& 4) on nuScenes validation set. Each row corresponds to the same object across time. We draw object detections in {\color{cyan}cyan}, \radar\ targets within past 0.5s in white, and associated \radar\ targets with $>0.1$ normalized score in {\color{Goldenrod}yellow}.}
\label{fig:attn}
\end{figure}

To better understand in which aspects the velocity estimation performance is improved by exploiting \radar\, we conduct fine-grained evaluation on the larger-scale DenseRadar dataset with respect to different subsets of object labels. In particular, we create different subsets of labels by varying the object distance to the ego vehicle, number of observed \lidar\ points, angle $\gamma$ between motion direction and radial direction, and the velocity magnitude.

We compare three model variants: \lidar\ only, our model with heuristic late fusion and our model in Fig. \ref{fig:ablation}.
From the results we see that the heuristic model brings negligible gains when $\gamma>\ang{10}$ or $\|\mathbf{v}\| < 3$ m/s. This justifies the $\ang{40}$ and $1$ m/s thresholds in our heuristics as these are cases where \radar\ data contain large uncertainty.
In contrast, our attention-based model consistently and significantly outperforms the heuristic model under all conditions, showing its effectiveness in capturing  sensor uncertainties and exploiting both sensors.

\subsection{Qualitative Results}
In Fig. \ref{fig:attn} we show the learned detection and \radar\ associations. Results are shown in sequence for each object to illustrate the temporal change in the association.
From the results we observe that: (1) the association is sparse in that only relevant \radar\ targets are associated; (2) the association is quite robust to noisy locations of the \radar\ targets; (3) the model captures the uncertainty of \radar\ targets very well. For example, when the radial direction is near tangential to the object's motion direction, the model tends to not associate any \radar\ targets as in such cases the \radar\ evidence is often very unreliable.

% !TEX root = top.tex
\section{Conclusion}
We have proposed a new method to exploit Radar  in combination with LiDAR for robust perception of dynamic objects in  self-driving.
To exploit  geometric information from Radar, we use a voxel-based early fusion approach, which is shown to improve long-distance object detection due to Radar's longer sensing range.
To exploit  dynamic information, we propose an attention-based late fusion approach, which addresses the critical problem of associating Radar targets and objects without ground-truth association labels.
By learning to associate and aggregate information, a significant performance boost in velocity estimation is observed under various conditions.

% \clearpage
\bibliographystyle{splncs04}
\bibliography{egbib}
\end{document}